\documentclass[journal,twoside]{IEEEtran}
\usepackage{amsmath,amsfonts}
\usepackage{algorithmic}
\usepackage{algorithm}
\usepackage{array}
\usepackage{textcomp}
\usepackage{stfloats}
\usepackage{url}
\usepackage{verbatim}
\usepackage{graphicx}
\usepackage{cite}

\usepackage{subfigure}
\usepackage{threeparttable}
\usepackage{makecell}
\usepackage{multirow}
\usepackage{color}
\usepackage{colortbl}
\usepackage{pifont}

\begin{document}

\title{Spectral-Spatial Global Graph Reasoning for Hyperspectral Image Classification}

\author{Di Wang,~\IEEEmembership{Member,~IEEE,}
        Bo Du,~\IEEEmembership{Senior Member,~IEEE,}
        Liangpei Zhang,~\IEEEmembership{Fellow,~IEEE} %and Yonghao~Xu,~\IEEEmembership{Student Member,~IEEE,}% <-this % stops a space

\thanks{D. Wang and B. Du are with the School of Computer Science, Wuhan University, Wuhan 430072, China (e-mail: wd74108520@gmail.com; dubo@whu.edu.cn).}
\thanks{L. Zhang is with the State Key Laboratory of Information Engineering in Surveying, Mapping and Remote Sensing, Wuhan University, Wuhan 430079, China (e-mail: zlp62@whu.edu.cn).}% <-this % stops a space
% <-this % stops a space
}

\markboth{Journal of \LaTeX\ Class Files,~Vol.~14, No.~8, August~2021}%
{Wang \MakeLowercase{\textit{et al.}}: SSGRN FOR HSI CLASSIFICATION}

%\IEEEpubid{0000--0000\copyright~2023 IEEE}

\maketitle

\begin{abstract}
  Convolutional neural networks have been widely applied to hyperspectral image classification. However, traditional convolutions can not effectively extract features for objects with irregular distributions. Recent methods attempt to address this issue by performing graph convolutions on spatial topologies, but fixed graph structures and local perceptions limit their performances. To tackle these problems, in this paper, different from previous approaches, we perform the superpixel generation on intermediate features during network training to adaptively produce homogeneous regions, obtain graph structures, and further generate spatial descriptors, which are served as graph nodes. Besides spatial objects, we also explore the graph relationships between channels by reasonably aggregating channels to generate spectral descriptors. The adjacent matrices in these graph convolutions are obtained by considering the relationships among all descriptors to realize global perceptions. By combining the extracted spatial and spectral graph features, we finally obtain a spectral-spatial graph reasoning network (SSGRN). The spatial and spectral parts of SSGRN are separately called spatial and spectral graph reasoning subnetworks. Comprehensive experiments on four public datasets demonstrate the competitiveness of the proposed methods compared with other state-of-the-art graph convolution-based approaches.
\end{abstract}


\section{Introduction}

\IEEEPARstart{R}{e}lying on excellent characteristics that intrinsic properties of targets can be identified by automatically extracting effective features in an end-to-end manner. Deep learning technologies are being extensively employed in the processing of hyperspectral images (HSIs). HSI includes abundant spectral information which is carried by hundreds of bands and vision representations presented by high spatial resolution, effectively serving precision agriculture \cite{agriculture_1}, environmental monitoring \cite{env_monit_1}, anomaly detection \cite{anomaly_1} and so on. Among these fields, hyperspectral image classification (HSIC) is always a fundamental and hot topic, where each pixel in the whole scene is needed to be assigned a unique semantic category.

Among existing deep learning technologies, convolutional neural network (CNN) is the most commonly used framework in the HSIC community \cite{hu2015deep, Zhao2016,3dcnn,ssrn,s3fse_tcyb,gllr_tcyb}. However, CNN can only aggregate the contexts in regular regions, while the objects in HSI usually have irregular distributions. Recently, many graph convolution network (GCN) \cite{kipfgcn} based methods are developed \cite{mdgcn,cadgcn,dasgcn,cegcn,digcn,emsgcn,gcgcn,dmsger,s3fgcn} to address this issue by treating objects as graph nodes. The graph structures are usually obtained through superpixel segmentation on the original image. However, the obtained fixed graph topologies still limit the performance.

Furthermore, conventional convolutions are local operators. It is difficult to model the dependencies between long-range positions, causing networks cannot fully leverage contextual information. For this problem, \cite{non-local} introduces a self-attention mechanism to capture non-local contexts. However, since the similarities between each pixel and all the other positions need to be computed, high computational costs are required.

To tackle the above problems, in this paper, different from previous approaches, our method can produce dynamic graph structures through more flexible homogeneous areas. These areas are generated by conducting superpixel segmentation on intermediate features inside networks based on spectral-spatial similarities between pixels. Then, the graph nodes are produced by these areas. These nodes are also called descriptors since each vector is obtained by aggregating the pixel representations of an area. Compared to the nodes of previous approaches, our descriptors are more discriminative since they are adaptively obtained from constantly changed homogeneous areas with network learning. In addition, the number of descriptors is usually much less than the pixels. Therefore, our method requires lower complexities compared with the aforementioned non-local modules. What's more, for the issue of local perception, the graph convolution in the proposed method is implemented in a global view to acquire more effective graph contexts. This can be implemented with the help of the self-attention mechanism. After the graph convolution, we obtain pixel-level results used for final classification by reasonably combining these descriptors.

It should be noticed that different from rich and complex natural image databases that have certain channels, HSI scene usually involves a single image with hundreds of bands that are determined by the sensor type. Thus, one of the key parts in HSI processing that can be distinguished from the natural image operation is to explore how to better exploit the spectral information. Existing literatures \cite{ssun,assmn} show that the adjacent bands of HSI also contain contextual information. Therefore, besides the graph convolution in the spatial aspect, we adopt a similar idea in the spectral aspect, i.e., reasonably aggregating these channels to generate spectral descriptors, and additionally employing graph convolution onto them to capture the relationships between different bands. This graph convolution is also implemented in a global perception manner.

Since our graph convolutions are both conducted globally, this is different from the general practice used in previous methods, where the convolution operation is only implemented on adjacent nodes. We call our method spectral-spatial graph reasoning network (SSGRN), which includes spatial and spectral two parts that are separately named spatial graph reasoning subnetwork (SAGRN) and spectral graph reasoning subnetwork (SEGRN). The main contributions of this paper can be summarized as follows:

\begin{itemize}
  \item[1)] We propose an end-to-end spectral-spatial graph reasoning network named SSGRN. Compared with existing spectral-spatial joint networks, our model can adaptively capture the contexts lying in different objects or channels, despite they are in irregular distributions.
  \item[2)] We design a spatial subnetwork called SAGRN, where the superpixel segmentation is trainable, adaptively generating flexible homogeneous areas to produce effective descriptors, and we perceive the relationships between any descriptors by adopting global graph reasoning.
  \item[3)] A spectral subnetwork SEGRN is proposed to capture the contextual information lying in different bands using graph reasoning. As far as we know, it is the first time the relationships of spectral channels are explored from a graph perspective for HSIC. 
  \item[4)] Benefitting from the proposed graph reasoning modules. Our networks achieve promising results on four HSIC benchmarks including Indian Pines, Pavia University, Salinas Valley, and University of Houston, compared with other GCN-based advanced methods.
\end{itemize}

The remainder of this paper is organized as follows. Section II gives an introduction of related works. Section III describes the proposed networks. Experiments and related comprehensive analyses are presented in section IV. Finally, Section V concludes the paper.

\section{Related Work}

In this section, we first introduce the history of deep learning-based methods for HSIC. Then, since the proposed method is a segmentation network involving graph convolution, we review the deep learning-related segmentation approaches. Finally, we present the approaches related to graph convolution in the HSIC field.
  
\subsection{Deep Learning Method for HSIC}

Early HSIC community extracts deep features by fully connected networks \cite{sae,dbn,zhoudsae2019}. However, these networks require sufficient computations since each neuron needs to connect with all units in the next layer. To tackle these problems, many researchers use CNNs for pixel-level classification. In addition, compared with the DNNs receiving 1-D spectral vectors, CNNs process the clipped spatial patches around target pixels and have larger vision fields. In the past years, regarding abundant channels, mainstream deep learning-based methods are classification networks that receive spatial patches or spectral vectors of target pixels \cite{hu2015deep, Zhao2016,3dcnn,ssrn,s3fse_tcyb,gllr_tcyb}. In addition to being directly used as an extractor for single-scale features, CNNs can be modified to generate enhanced features for further accuracy improvement \cite{rpnet,ssun,assmn,ssan,acd_2021_tgrs_dsamnet,specattennet,3dadnet,ssatt,docssan,HE2022102667,rssan,Lee2017TIP,zhang2018tipforHSIC,gonghier2018}. However, as networks deepen, their abilities are soon saturated, because of the limited sizes of input patches despite theoretical receptive field may extraordinarily large. To this end, many methods adopt the fully convolutional network (FCN) --- a special family of CNNs, where the whole image can be directly input into the network and all pixels are simultaneously classified \cite{ssfcn,freenet,fcontnet,ptop_cnn_tcyb}. Although CNN has made promising achievements in HSIC, traditional convolutions still cannot meet the requirements of obtaining the relationships between objects with irregular distributions. Therefore, GCN-based methods are being developed. We will introduce the related technologies later.

\subsection{Semantic Segmentation}

Most existing deep learning-based segmentation methods adopt FCN \cite{fcn_cvpr}. For example, UNet \cite{unet} and RefineNet \cite{refinenet} use encoder-decoder architecture to carefully recover details for upsampling. However, simple stacking ordinary $3 \times 3$ convolution causes limited perceptions. To expand the vision field, GCN (global convolution network) \cite{globalconv} uses larger kernels, BiSeNet \cite{bisenet} adopts global pooling, while an effective encoding layer is introduced in EncNet \cite{encnet}. 
Multiscale features are employed to improve segmentation performance, such as Deeplab \cite{deeplabv3_arxiv} and PSPNet \cite{pspnet}, which separately benefit from ASPP or PPM modules that adopt multiple dilated convolutions or spatial poolings in different scales. With flexible long-distance perception, self-attention mechanisms have also been introduced into the segmentation community. For instance, DANet \cite{danet} utilizes position and channel attention to separately obtain spatial and channel contexts. Pyramid-OCNet \cite{ocnet} restricts the spatial context capturing in grids generated by the PPM module. While in CCNet \cite{ccnet}, only pixels lying in a criss-cross area can communicate with each other. Recently, descriptor-based methods are gradually being valued by researchers. To obtain descriptors, ACFNet \cite{acfnet} aggregates features of the same category in the label map, OCRNet \cite{ocrnet} adopts the generated soft regions with the help of an auxiliary loss branch, while EMANet \cite{emanet} directly defines and learns descriptors by Expectation-Maximization algorithm. In addition, the descriptors can interact with each other in various ways, and a typical implementation is with graph convolution \cite{glore}, whose applications in the HSIC field will be detailed later.

In the HSIC domain, the development history of using segmentation networks is similar to the natural image. Nevertheless, there are still some differences because of the distinct characteristics of HSI. The earliest SSFCN \cite{ssfcn} applies dilated convolutions into spatial and spectral feature extractions for image-level classification. Although FPGA \cite{freenet} also adopts an encoder-decoder structure, it extra designs a stochastic stratified sampler to promote network convergence. ENL-FCN \cite{enl_fcn} and FullyContNet \cite{fullycontnet} employ self-attention mechanisms, too, and the latter captures sufficient contextual information by simultaneously introducing spatial, channel, and scale attentions. Unlike them, the proposed methods follow the pattern of employing descriptors to save computational overhead. Concretely, our SAGRN obtains the descriptors by gathering features in homogeneous areas that are generated by superpixel segmentation inside networks, while the counterparts in SEGRN are obtained by aggregating closely related adjacent channels.

\subsection{HSIC Using GCN}

According to our literature survey, GCN is being widely used in the HSIC community, and many approaches \cite{mdgcn,cadgcn,dasgcn,cegcn,digcn,emsgcn,gcgcn,dmsger,s3fgcn} are constructed based on superpixel segmentation, since it can provide natural graph structure. For example, MDGCN \cite{mdgcn} generates descriptors by directly superpixel segmenting the original image, and then performs the superpixel-level labeling. \cite{cadgcn} initializes graph nodes with superpixel centers, and then the nodes are adjusted by modeling the relationships between the center and the pixels within a superpixel, the final classification map is obtained through a region-pixel transformation. Besides using a predefined graph. DGCN \cite{dgcn} separately generates point and distribution graphs with deep features to optimize the relationships between training samples, while EMS-GCN \cite{emsgcn} conducts the superpixel segmentation procedure inside the network to generate an adaptive graph, too. Similar to multiple conventional convolutions, DIGCN \cite{digcn} and XGPN \cite{xgpn} adopt parallel graph convolutions in different scales. While CEGCN \cite{cegcn} and LA-DG-GCN \cite{ladggcn} further combine the features from GCN and CNN, respectively. Besides the spatial aspect, spectral information is also drawing researchers' attention, and it can be exemplified by DASGCN \cite{dasgcn}. In addition, some approaches aim to simplify the calculation of the original GCN for efficient HSIC. For instance, \cite{minigcn} operates on smaller graphs using mini-batch samples and the extracted sub-adjacent matrices. Some other schemes, such as spatial pooling \cite{spgcn}, novel distance metric \cite{metric_gcn} and position coordinates \cite{coordin_gcn} are being involved in the graph convolution procedure to achieve HSIC.

Compared with the above attempts, we must argue that the proposed methods are different from them. Firstly, in SEGRN, a spectral graph is adaptively generated since the nodes are obtained by aggregating adjacent channels of changing intermediate features in the network, while the corresponding graph in DASGCN is predefined. What’s more, although the descriptors in SAGRN are also built by superpixel segmentation on the intermediate feature, our spatial graph is dense, since our adjacent matrix is obtained by modeling the relationships between any descriptor with all other partners. Thus, our descriptors can capture global contexts, which are completely different from EMS-GCN, whose adjacent matrix is still produced through traditional similarity computation between neighboring nodes. At last, the proposed SSGRN can adaptively obtain spectral and spatial global graph contextual information.

\section{Proposed Methods}
In this section, we first briefly introduce the definition of GCN. Then, the proposed networks including SAGRN, SEGRN, and SSGRN will be successively presented.

\subsection{Graph Convolutional Network}

The original GCN \cite{kipfgcn} is defined as follows
\begin{equation}
  \mathcal{G}^{(l)} = \left\{
    \begin{array}{ll}
    X,\quad &l=0\\
    \sigma\left(Z\mathcal{G}^{(l-1)}W^{(l)} \right),\quad &l>0
  \end{array}
  \right.
\end{equation}
Here, $X$ is the input feature, $Z= \widehat{D}^{-\frac{1}{2}} \widehat{A} \widehat{D}^{-\frac{1}{2}}$, where $\widehat{D}_{ii}=\sum_{j=1}^K \widehat{A}_{ij}, \widehat{A}=\widetilde{A}+I_K$. $\widetilde{A}$ is the adjacency matrix of the current graph that has $K$ nodes, and $I_K$ is the identity matrix, $\mathcal{G}^{(l)}$ and $W^{(l)}$ are separately the output and trainable parameter matrices of the $l$th layer, $\sigma$ is the ReLU activation function.

From the above formula, we can see that the critical parts for graph reasoning are to effectively obtain $Z$ and $X$, while the proposed networks realize similar ideas.

\begin{figure*}[t]
  \centering
  \includegraphics[width=\linewidth]{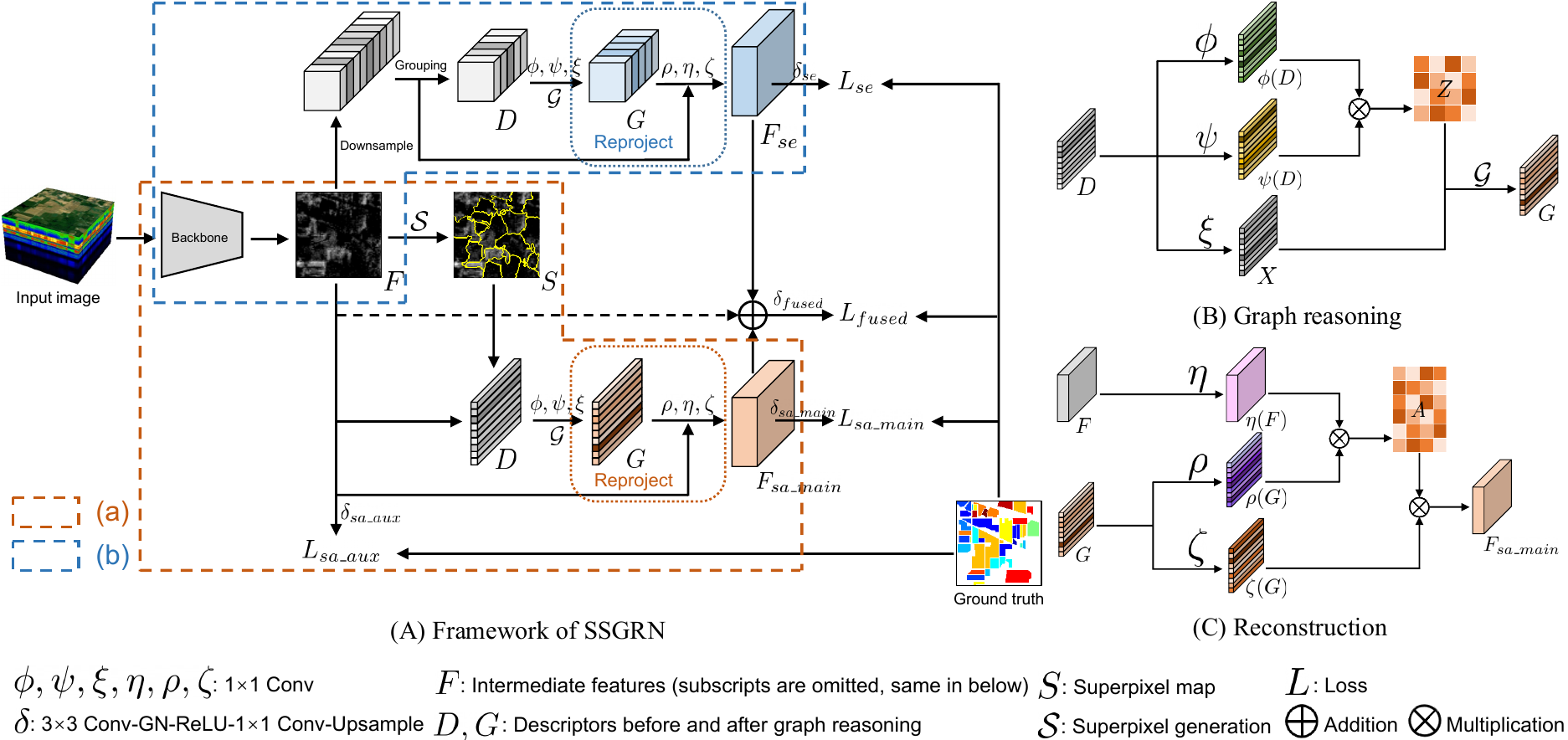}
  \caption{The architecture of the proposed method. (A) shows the Framework of the SSGRN, where feature $F$ is firstly obtained after the input image through the backbone network. Then $F$ is separately fed into two parts of (a) SAGRN and (b) SEGRN. In the spatial part, a superpixel map $S$ is generated based on $F$ and used to produce a descriptor set $D$. Graph reasoning is then conducted on $D$, and the generated $G$ is employed to reconstruct pixel-level feature $F_{sa\_main}$ for classification under the guidance of $F$.  An auxiliary branch (AB) is imposed on $F$ to improve the quality of $S$. In the spectral subnetwork, to obtain pixel-level feature $F_{se}$, grouping, graph reasoning, and reconstruction are sequentially implemented on the foundation of features that are obtained by downsampling $F$. In the end, a skip connection is built on $F$ to aggregate with $F_{sa\_main}$ and $F_{se}$ for final classification. (B) and (C) present detailed graph reasoning and reprojection pipelines of SAGRN, respectively. Except for downsampling, SEGRN has similar procedures.}
  \label{network}
\end{figure*}

\subsection{Spatial Graph Reasoning Subnetwork}

Hundreds of bands and high spatial resolution in HSI provide a close spectral-spatial relationship between pixels, and this contextual information is easily utilized by superpixel segmentation, which can be implemented by adopting the SLIC algorithm \cite{slic} to generate a series of compact superpixels. However, the original SLIC algorithm is difficult to be directly placed into the network for end-to-end training because of undifferentiable min-max operations. \cite{ssn_eccv} addresses this problem by transforming these operations to differentiable weighted addition. With this technique, we successfully obtain superpixel segments inside the network and generate effective descriptors. In addition, it should be noticed that the generated superpixels bring graph structure and the number of clustered districts is significantly less than the original pixels, benefitting conducting graph reasoning with high efficiency.

In SAGRN, the obtained superpixels and descriptors are actually homogeneous regions and a group of vectors $D=\left\{d_1,d_2,\cdots, d_K\right\}\in \mathbb{R}^{K \times C}$. Concretely, each descriptor is computed by taking the average of the features in the corresponding area. Then, the input $X=\left\{x_1,x_2,\cdots, x_K\right\}$ of the GCN is obtained from $D$ using linear mapping. This process is shown as follows
\begin{equation}
  \begin{split}
  S & = \mathcal{S}(F)\\
  d_i & = \frac{\sum_{j=1}^{HW}\mathbb{I}(S_j=i)\cdot F_j}{\sum_{j=1}^{HW}\mathbb{I}(S_j=i)} \quad i=1,\cdots,K\\
  x_i & = \xi\left(d_i\right)
  \end{split}
\end{equation}
where $F\in \mathbb{R}^{C\times H\times W}$ is the input feature, $\mathcal{S}$ and $S$ represent the superpixel generating procedure and the corresponding segmentation result map. There are a total of $K$ descriptors and $i$ is the index, $C$, $H$ and $W$ are separately the number of channels, height, and width of $F$. $\mathbb{I}(S_j=i)$ is a binary indicator that judges whether the value of the $j$th pixel in $S$ is equal to $i$. $\xi$ represents a 1 $\times$ 1 convolutional layer, which is used to conduct information integration for different channels.

To obtain $Z$, different from the conventional adjacent matrix that only considers neighbor nodes, we treat it as a dense graph where each node possesses a relationship with all the other nodes to better capture graph contexts. Specifically, the relationship in each pair of nodes is measured by computing their similarity in a mapped latent space
\begin{equation}
  Z_{ij}=\frac{\exp(\phi(d_i)^T\psi(d_j))}{\sum_{k=1}^K \exp(\phi(d_i)^T\psi(d_k))} \quad i=1,\cdots,K
\end{equation}
where $\phi$ and $\psi$ are mapping functions, both of which can be implemented with a 1 $\times$ 1 convolution. Then, we normalize the similarity matrix using the softmax function.

After obtaining $Z \in \mathbb{R}^{K \times K}$ and $X\in \mathbb{R}^{K \times C_1}$, spatial graph reasoning is achieved by directly adopting the GCN formula
\begin{equation}
  \mathcal{G}(Z,X)=\sigma\left(ZXW\right)
\end{equation}
where $W\in \mathbb{R}^{C_1 \times C_2}$ is trainable parameters and ReLU is used as the activation function $\sigma$. It can be seen that each district-level feature $x$ is enhanced since it has a global view that captures the contextual information lying in all the other nodes. 

At last, these enhanced nodes need to be reprojected for recovering the shape of pixel-level features. The construction of pixel-level features depends on a reasonable combination of descriptors since they possess specific connotations that are various from each other. In this paper, the node vectors after reasoning are considered as a group of bases that can form an effective feature space, where the information at any point can be inferred based on a linear aggregation of these vectors for more complex semantic understanding. 

To this end, assume $G = \mathcal{G}(Z,X)=\left\{g_1,g_2,\cdots,g_K\right\} \in \mathbb{R}^{K \times C_2}, F=\left\{f_1,f_2,\cdots,f_{HW}\right\}$. The affinities $A \in \mathbb{R}^{K \times HW}$ between feature $F$ and node set $G$ are firstly measured in a newly transformed space
\begin{equation}
  A_{ij}=\frac{\exp(\rho(g_i)^T\eta(f_j))}{\sum_{h=1}^{HW} \exp(\rho(g_i)^T\eta(f_h))}  \quad i=1,\cdots,K
\end{equation}
Then the target pixel-level feature $F_{sa\_main} \in \mathbb{R}^{HW \times C_3}$ is subsequently obtained by linearly combining these graph nodes, where the affinity matrix $A$ is served as corresponding weights. Thus, $F_{sa\_main}=A^T \zeta(G)$, and we subsequently reshape $F_{sa\_main}$ to $\mathbb{R}^{C_3 \times H\times W}$. Here, $\rho$, $\eta$, and $\zeta$ are all implemented with a 1 $\times$ 1 convolution, and the subscript \textit{main} means the main branch to distinguish the later introduced auxiliary path in the network. In the above procedure, $C=C_1=C_2=C_3$ for convenience. After obtaining the $F_{sa\_main}$, through a series of layers, denoted as $\delta_{sa\_main}$, including a 3 $\times$ 3 convolutional layer followed by a group normalization (GN) layer, a ReLU function, a 1 $\times$ 1 convolutional layer, and a bilinear upsampling function, the probability matrix $P_{sa\_main}\in \mathbb{R}^{C_n \times H \times W}$ is acquired for computing loss, where $C_n$ is the number of categories. This procedure is symbolized as follows.
\begin{equation}
  P_{sa\_main} = \delta_{sa\_main}(F_{sa\_main})
\end{equation}

However, in the early stage of training, disorganized high-level features may be unfavorable to the homogeneous region generation, and classification quality is unavoidably suffered with the affected descriptors. To produce more stable superpixels, we add an auxiliary branch (AB) to achieve fast convergence and obtain probability matrix $P_{sa\_aux}$ like the main branch.
\begin{equation}
  P_{sa\_aux} = \delta_{sa\_aux}(F)
\end{equation}

In the proposed methods, the loss function is defined to $L_{(\cdot)}=l(P_{(\cdot)}, Y)$, where $Y$ is the ground truth and $l$ is implemented with the cross-entropy loss, thus the total loss of SAGRN is
\begin{equation}
  L_{sa} = L_{sa\_main} + L_{sa\_aux}
\end{equation}

The diagram of SAGRN is presented in Figure \ref{network} (A)-(a).

\subsection{Spectral Graph Reasoning Subnetwork}

In SEGRN, we adopt a similar idea as SAGRN. Since the homogeneous areas in SAGRN are regarded as clusters of pixels. Thus, it is natural to consider reasonably obtaining channel clusters. For this purpose, inspired by \cite{ssun,assmn}, we directly take the mean value of adjacent channels as spectral descriptors. This can be realized by grouping the feature maps in the channel direction.

Concretely, for a input feature $F'=\left\{b_1,b_2,\cdots,b_C\right\} \in \mathbb{R}^{C \times H'\times W'}$ with $C$ bands, assume they are separately assigned to $M$ groups $F'=\left\{r_1,r_2,\cdots,r_M\right\}$, then the $i$th group is
\begin{equation}
  r_i = \left\{b_{\frac{C}{M}(i-1)+1},b_{\frac{C}{M}(i-1)+2},\cdots,b_{\frac{C}{M} \cdot i}\right\}
\end{equation}
Thus the $i$th spectral descriptor can be obtained through
\begin{equation}
  d_i = \frac{\sum_{j=1}^{C/M} b_{\frac{C}{M}(i-1)+j}}{C/M}
\end{equation}

The remaining steps are similar to SAGRN. It should be noticed that $F'$ is downsampled from $F$ by average pooling before conducting graph reasoning to save computational resources. Therefore, $H'<H$, $W'<W$, and we group the downsampled $F'$ to generate spectral descriptors $d$. The pixel-level feature $F_{se}\in \mathbb{R}^{C\times H\times W}$ is reconstructed after a reasonable descriptor linear combination. Then, a bilinear interpolation is employed to ensure the feature sizes are consistent in subsequent aggregations. Through SEGRN, we successfully perform graph reasoning in spectral direction since the contextual information lying in different channels is perceived, and we obtain an enhanced feature, where each channel is improved by capturing the contexts of other bands. 

At last, the loss of SEGRN is computed by
\begin{equation}
  L_{se} = l\left(\delta_{se}(F_{se}),Y\right)
\end{equation}

The diagram of SEGRN is depicted in Figure \ref{network} (A)-(b).

\subsection{Spectral-Spatial Graph Reasoning Network}

The feature $F$ is obtained through a backbone network containing three blocks. Each block includes a convolutional layer followed by a GN layer and a ReLU function. There is a 2x downsampling after the first block to reduce memory consumption. The whole network is trained from scratch and does not need any pre-trained parameters of existing popular models.

After passing through SAGRN and SEGRN, we obtain the corresponding enhanced features $F_{sa\_main}$ and $F_{se}$. To preserve the original information of the input feature $F$, we adopt a residual skip connection, thus the spectral-spatial fused feature $F_{fused}\in \mathbb{R}^{C\times H\times W}$ is defined as
\begin{equation}
  F_{fused} = F_{sa\_main} + F_{se} + F
\end{equation}
And the corresponding loss is also obtained by
\begin{equation}
  L_{fused} = l\left(\delta_{fused}(F_{fused}),Y\right)
\end{equation}

At last, the total loss of the proposed SSGRN is computed through
\begin{equation}
  L_{ss} = L_{sa} + L_{se} + L_{fused}
\end{equation}
The whole diagram of SSGRN is shown in Figure \ref{network} (A).

\section{Experiments}
In this section, we first introduce the used datasets and implementation details, then we conduct a series of comprehensive assessments of the proposed methods, including hyperparameter selection, module ablation study, and model complexity analysis. The performance and stability comparisons between our methods with other state-of-the-art approaches are subsequently presented. In the end, we visualize some internal variables to further assist in understanding the mechanisms inside the network.

\subsection{Dataset}

\definecolor{16_1}{RGB}{0, 0, 191}
\definecolor{16_2}{RGB}{0, 0, 255}
\definecolor{16_3}{RGB}{0, 63, 255}
\definecolor{16_4}{RGB}{0, 127, 255}
\definecolor{16_5}{RGB}{0, 191, 255}
\definecolor{16_6}{RGB}{0, 255, 255}
\definecolor{16_7}{RGB}{63,255,191}
\definecolor{16_8}{RGB}{127,255,127}
\definecolor{16_9}{RGB}{191,255,63}
\definecolor{16_10}{RGB}{255,255,0}
\definecolor{16_11}{RGB}{255,191,0}
\definecolor{16_12}{RGB}{255,127,0}
\definecolor{16_13}{RGB}{255,63,0}
\definecolor{16_14}{RGB}{255,0,0}
\definecolor{16_15}{RGB}{191,0,0}
\definecolor{16_16}{RGB}{127,0,0}

\begin{table}[t]
\caption{Category and sample settings of the Indian Pines dataset}
\centering
\tiny
%\resizebox{\linewidth}{!}{
  \begin{tabular}{llccccc}
\hline
Class ID& Category & Color & Training  & Validation & Testing & Total \\
\hline
1 & Alfalfa & \cellcolor{16_1} & 26 & 7 & 13 & 46 \\
2 & Corn-notill & \cellcolor{16_2} & 80 &20 & 1328 & 1428  \\
3 & Corn-mintill& \cellcolor{16_3} & 80 & 20& 730 & 830  \\
4 & Corn & \cellcolor{16_4} & 80 &20 & 137 & 237  \\
5 & Grass-pasture & \cellcolor{16_5} & 80 &20 & 383 & 483  \\
6 & Grass-trees & \cellcolor{16_6} & 80 & 20& 630 & 730  \\
7 & Grass-pasture-mowed & \cellcolor{16_7} & 16 &4 & 8 & 28  \\
8 & Hay-windrowed & \cellcolor{16_8} & 80 &20 & 378 & 478  \\
9 & Oats & \cellcolor{16_9} & 11 &3 & 6 & 20 \\
10 & Soybean-notill & \cellcolor{16_10} & 80 & 20& 872 & 972  \\
11 & Soybean-mintill & \cellcolor{16_11} &80 &20 & 2355 & 2455  \\
12 & Soybean-clean & \cellcolor{16_12} & 80 &20 & 493 & 593 \\
13 & Wheat & \cellcolor{16_13} & 80 &20 & 105 & 205  \\
14 & Woods & \cellcolor{16_14} & 80 &20 & 1165 & 1265  \\
15 & Buildings-Grass-Trees-Drives & \cellcolor{16_15} & 80 &20 & 286 & 386  \\
16 & Stone-Steel-Towers & \cellcolor{16_16} & 60 &15 & 18 & 93  \\
\hline
Total &  & & 1073 & 269 & 8907 & 10249\\
\hline
\end{tabular}
%}
\label{indian_dataset_tab}
\end{table}

\definecolor{9_1}{RGB}{0, 0, 255}
\definecolor{9_2}{RGB}{0, 85, 255}
\definecolor{9_3}{RGB}{0, 170, 255}
\definecolor{9_4}{RGB}{0, 255, 255}
\definecolor{9_5}{RGB}{85,255, 170}
\definecolor{9_6}{RGB}{170,255,85}
\definecolor{9_7}{RGB}{255,255,0}
\definecolor{9_8}{RGB}{255,170,0}
\definecolor{9_9}{RGB}{255,85,0}

\begin{table}[t]
\caption{Category and sample settings of the Pavia University dataset}
\centering
\tiny
%\resizebox{\linewidth}{!}{
  \begin{tabular}{llccccc}
\hline
Class ID& Category & Color & Training  & Validation & Testing & Total\\
\hline
1 & Asphalt & \cellcolor{9_1} & 80 & 20 &  6531 & 6631 \\
2 & Meadows & \cellcolor{9_2} & 80 &20 & 18549 & 18649 \\
3 & Gravel & \cellcolor{9_3} & 80 & 20& 1999 & 2099 \\
4 & Trees & \cellcolor{9_4} & 80 &20 & 2964 & 3064 \\
5 & Metal sheets& \cellcolor{9_5} & 80 &20 & 1245 & 1345 \\
6 & Bare soil & \cellcolor{9_6}& 80 & 20& 4929 & 5029 \\
7 & Bitumen & \cellcolor{9_7}& 80 &20 & 1230 & 1330 \\
8 & Bricks & \cellcolor{9_8}& 80 &20 & 3582 & 3682 \\
9 & Shadows & \cellcolor{9_9} & 80 &20 & 847 & 947 \\
\hline
Total &  & & 720 & 180 & 41876 & 42776\\
\hline
\end{tabular}
\label{pavia_dataset_tab}
%}
\end{table}

\begin{table}[t]
      
      \label{indian}
      \caption{Category and sample settings of the Salinas Valley dataset}
      \centering
      \tiny
      %\resizebox{\linewidth}{!}{
        \begin{tabular}{llccccc}
      \hline
      Class ID& Category & Color & Training  & Validation & Testing & Total\\
      \hline
      1 & Brocoli green weeds 1 & \cellcolor{16_1} & 80 & 20 & 1909 & 2009 \\
      2 & Brocoli green weeds 2 & \cellcolor{16_2} & 80 &20 & 3626 & 3726 \\
      3 & Fallow & \cellcolor{16_3} & 80 & 20& 1876 & 1976 \\
      4 & Fallow rough plow & \cellcolor{16_4} & 80 &20 & 1294 & 1394 \\
      5 & Fallow smooth & \cellcolor{16_5} & 80 &20 & 2578 & 2678 \\
      6 & Stubble & \cellcolor{16_6} & 80 & 20& 3859 & 3959 \\
      7 & Celery & \cellcolor{16_7} & 80 &20 & 3479 & 3579 \\
      8 & Grapes untrained & \cellcolor{16_8} & 80 &20 & 11171 & 11271 \\
      9 & Soil vinyard develop & \cellcolor{16_9} & 80 &20 & 6103 & 6203 \\
      10 &Corn senesced green weeds & \cellcolor{16_10} & 80 & 20& 3178 & 3278 \\
      11 &Lettuce romaine 4wk & \cellcolor{16_11} &80 &20 & 968 & 1068 \\
      12 &Lettuce romaine 5wk & \cellcolor{16_12}  & 80 &20 & 1827 & 1927 \\
      13 &Lettuce romaine 6wk & \cellcolor{16_13}  & 80 &20 & 816 & 916 \\
      14 &Lettuce romaine 7wk & \cellcolor{16_14} & 80 &20 & 970 & 1070 \\
      15 &Vinyard untrained & \cellcolor{16_15} & 80 &20 & 7168 & 7268 \\
      16 &Vinyard vertical trellis & \cellcolor{16_16} & 80 &20 & 1707 & 1807 \\
      \hline
      Total&    &  & 1280 & 320 & 52529 & 54129\\
      \hline
    \end{tabular}
    %}
    \label{salina_dataset_tab}
      \end{table}

\definecolor{15_1}{RGB}{0, 0, 191}
\definecolor{15_2}{RGB}{0, 0, 255}
\definecolor{15_3}{RGB}{0, 63, 255}
\definecolor{15_4}{RGB}{0, 127, 255}
\definecolor{15_5}{RGB}{0, 191, 255}
\definecolor{15_6}{RGB}{0, 255, 255}
\definecolor{15_7}{RGB}{63,255,191}
\definecolor{15_8}{RGB}{127,255,127}
\definecolor{15_9}{RGB}{191,255,63}
\definecolor{15_10}{RGB}{255,255,0}
\definecolor{15_11}{RGB}{255,191,0}
\definecolor{15_12}{RGB}{255,127,0}
\definecolor{15_13}{RGB}{255,63,0}
\definecolor{15_14}{RGB}{255,0,0}
\definecolor{15_15}{RGB}{191,0,0}

\begin{table}[t]
 
  \caption{Category and sample settings of the University of Houston dataset}
  \centering
  \tiny
  %\resizebox{\linewidth}{!}{
  \begin{tabular}{llccccc}
  \hline
  Class ID& Category &Color & Training  & Validation & Testing & Total\\
  \hline
  1 & Grass healthy & \cellcolor{15_1} & 80 & 20 &  1151 & 1251 \\
  2 & Grass stressed & \cellcolor{15_2} & 80 &20 & 1154 & 1254 \\
  3 & Grass synthetic & \cellcolor{15_3} & 80 & 20& 597 & 697 \\
  4 & Trees & \cellcolor{15_4} & 80 &20 & 1144 & 1244 \\
  5 & Soil & \cellcolor{15_5} & 80 &20 & 1142 & 1242 \\
  6 & Water & \cellcolor{15_6} & 80 & 20& 225 & 325 \\
  7 & Residential & \cellcolor{15_7} & 80 &20 & 1168 & 1268 \\
  8 & Commercial & \cellcolor{15_8} & 80 &20 & 1144 & 1244 \\
  9 & Road & \cellcolor{15_9} & 80 &20 & 1152 & 1252 \\
  10 & Highway & \cellcolor{15_10} & 80 & 20& 1127 & 1227 \\
  11 & Railway & \cellcolor{15_11} &80 &20 & 1135 & 1235 \\
  12 & Parking lot1 & \cellcolor{15_12} & 80 &20 & 1134 & 1234 \\
  13 & Parking lot2 & \cellcolor{15_13} & 80 &20 & 369 & 469 \\
  14 & Tennis court & \cellcolor{15_14} & 80 &20 & 328 & 428 \\
  15 & Running track & \cellcolor{15_15} & 80 &20 & 560 & 660 \\
  \hline
  Total &  & & 1200 & 300 & 13511 & 15011\\
  \hline
\end{tabular}
%}

\label{houston_dataset_tab}
  \end{table}

  \begin{itemize}
    \item[1)] Indian Pines: This scene was gathered in North-western Indiana by the Airborne Visible/Infrared Imaging Spectrometer (AVIRIS) sensor in 1992, consisting 200 bands with the size of 145 $\times$ 145 pixels that are in 20m spatial resolution after water absorption bands were removed and in the wavelength range of 0.4-2.5$\mu$m. 16 vegetation classes are involved in this scene. The settings of category color, training, validation, and testing samples have been presented in Table \ref{indian_dataset_tab}.
    \item[2)] Pavia University: This scene was obtained over Pavia University in Northern Italy by Reflective Optics System Imaging Spectrometer (ROSIS) in 2001, consisting 103 bands with the size of 610 $\times$ 340 pixels that are in 1.3m spatial resolution and the wavelength range of 0.43-0.86$\mu$m. 9 categories are included in this data. The experimental settings are shown in Table \ref{pavia_dataset_tab}.
    \item[3)] Salinas Valley: This scene was collected over Salinas Valley, California by the AVIRIS sensor, consisting 204 bands with the size of 512 $\times$ 217 pixels in 3.7m spatial resolution. Same as the Indian Pines scene, 20 water absorption bands were discarded. 16 categories including vegetables, bare soils, and vineyard fields are involved, see Table \ref{salina_dataset_tab} for experimental settings. 
    \item[4)] University of Houston: This scene was acquired over the University of Houston campus and its neighboring regions by ITRES-CASI 1500 sensor in 2012, containing 144 bands with the size of 349 $\times$ 1905 pixels that in 2.5m spatial resolution and in the wavelength range of 0.4-1.0$\mu$m. 15 categories are included in this data. Table \ref{houston_dataset_tab} lists the experimental configurations. This scene was also used in the 2013 IEEE GRSS Data Fusion Contest.
    \end{itemize}

\subsection{Implementation Details and Experimental Settings}

\begin{table}[t]
  \scriptsize
  \caption{Parameter details of the proposed methods}
  \newcommand{\tabincell}[2]{\begin{tabular}{@{}#1@{}}#2\end{tabular}}
  \centering
  \label{netdetail}
  \begin{threeparttable}
  %\resizebox{0.9\linewidth}{!}{
  \begin{tabular}{l|ccc}
  \hline
  Part & Layer & Parameter & GN \& ReLU\\
  \hline
  \multirow{4}{*}{Backbone}& Block1 & 64, 3, 1, 1, 1 &  \ding{52}     \\
  \cline{2-4}
                           & Max Pool & 64, 2, 2, 0, 1&        \\
  \cline{2-4}
                           & Block2 & 128, 3, 1, 1, 1 &  \ding{52}       \\
  \cline{2-4}
                           & Block3 & 256, 3, 1, 1, 1 &  \ding{52}       \\
\hline
\multirow{8}{*}{SAGRN}& \tnote{1} $\phi$ & 64, 1, 1, 0, 1 & \\
\cline{2-4}
                      & $\psi$ & 64, 1, 1, 0, 1 & \\
\cline{2-4}
                      & $\xi$  & 16, 1, 1, 0, 1&  \\
\cline{2-4}
                      & $\rho$ & 64, 1, 1, 0, 1 & \\
\cline{2-4}
                      & $\eta$ & 64, 1, 1, 0, 1 & \\
\cline{2-4}
                      & $\zeta$ & 16, 1, 1, 0, 1&  \\
\cline{2-4}
                      & $\delta_{sa\_aux}$ & \tabincell{c}{128, 1, 1, 0, 1 \\ \tnote{2} $C_n$, 1, 1, 0, 1} & \tabincell{c}{\ding{52} \\ \hspace{0.01mm}}\\
\cline{2-4}
                      & $\delta_{sa\_main}$ & \tabincell{c}{128, 1, 1, 0, 1 \\ $C_n$, 1, 1, 0, 1} & \tabincell{c}{\ding{52} \\ \hspace{0.01mm}}\\
\hline
\multirow{7}{*}{SEGRN}& $\phi$ & \tnote{3} $HW/64 $, 1, 1, 0, 1 & \\
\cline{2-4}
                      & $\psi$ & $HW/64$, 1, 1, 0, 1 & \\
\cline{2-4}
                      & $\xi$ & $HW/16$, 1, 1, 0, 1 &  \\
\cline{2-4}
                      & $\rho$ & $HW/64$, 1, 1, 0, 1 & \\
\cline{2-4}
                      & $\eta$ & $HW/64$, 1, 1, 0, 1 & \\
\cline{2-4}
                      & $\zeta$ & $HW/16$, 1, 1, 0, 1 &  \\
\cline{2-4}
                      & $\delta_{se}$ & \tabincell{c}{128, 1, 1, 0, 1 \\ $C_n$, 1, 1, 0, 1} & \tabincell{c}{\ding{52} \\ \hspace{0.01mm}}\\
\hline
SSGRN & $\delta_{fused}$ & \tabincell{c}{128, 1, 1, 0, 1 \\ $C_n$, 1, 1, 0, 1} & \tabincell{c}{\ding{52} \\ \hspace{0.01mm}} \\
\hline

\end{tabular}
%}
\begin{tablenotes}
  %\scriptsize
  \item[1] Sub scripts are omitted to align with Figure \ref{network}.
  \item[2] Category number.
  \item[3] $H$ and $W$ are height and width of the feature $F$.
\end{tablenotes}
\end{threeparttable}
\end{table}

\begin{figure}[t]
  \centering
  \includegraphics[width=\linewidth]{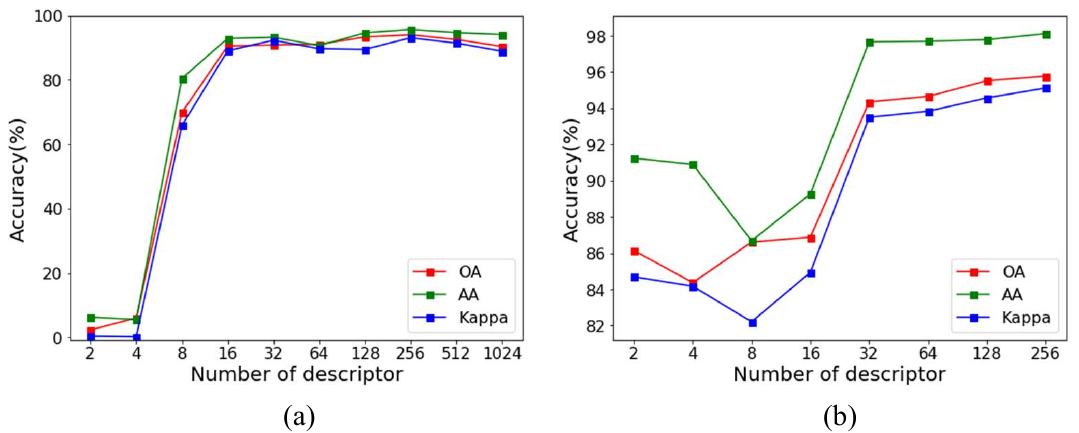}
  \caption{Accuracies versus the descriptor number on the Indian Pines dataset. (a) SAGRN. (b) SEGRN.}
  \label{k_acc}
\end{figure}

We employ Pytorch to implement the proposed methods. The base learning rate is set to 1e-3, which is adaptively adjusted with the poly scheduling strategy and multiplied by $\left(1-\frac{iter}{max\_iter}\right)^{0.9}$. The SGD with momentum optimization algorithm is used for training, where the momentum is set to 0.9 and the weight decay is configured as 0.0001. During training, iteration and batch size are always 1 in each epoch since the whole image is directly input into the network, and the total training iteration number is set to 1K. The structure details of the proposed networks are shown in Table \ref{netdetail} and the values of ``Parameter'' separately represent ``channel number'', ``kernel size'', ``stride'', ``padding size'' and ``dilation rate'' of the corresponding layer. In addition, this table also indicates whether GN and ReLU are used after the convolutional layer. Besides standardization, we do not adopt any data augmentations on the proposed methods.

In our implementations, training samples are obtained by randomly choosing. The remainder consists of a validation set and a testing set. Accuracies are evaluated on the testing set, while the validation set is used for monitoring model status during training. All experiments are repeated 10 times. Three commonly used evaluation criteria in the HSIC community are applied in the experiments, including overall accuracy (OA), average accuracy (AA), and Kappa coefficient (Kappa). OA is the most popular evaluation criterion, which is calculated by dividing the number of correctly classified pixels by the number of pixels that need to be judged. However, OA is usually affected by the phenomenon of unbalance categories. To tackle this problem, AA and Kappa are separately computed based on the confusion matrix. The recall values of all categories are averaged to get AA, while the Kappa is used for measuring classification consistency to penalize the model possessing category preference. All experiments are conducted using an NVIDIA Tesla V100 GPU.

\begin{table*}[t]
  \scriptsize
  \caption{Accuracies with various settings of the proposed methods on the Indian Pines dataset}
  \newcommand{\tabincell}[2]{\begin{tabular}{@{}#1@{}}#2\end{tabular}}
  \centering
  \label{ablation}
  %\resizebox{\linewidth}{!}{
    \begin{tabular}{l|ccccccc}
  \hline
  Method & SAGR & SEGR & \tabincell{c}{Auxiliary\\ Branch} & \tabincell{c}{Multiple \\ Loss} & OA(\%) & AA(\%) & Kappa(\%) \\
  \hline
  FCN &   &    &     &    /  &  91.06 &  96.35 &  89.71  \\
  \hline
  FCN+SAGR & \ding{52} & /  &     &  /& 89.87 & 91.81  &   88.31   \\
  SAGRN (FCN+SAGR+AB) & \ding{52} &  / &  \ding{52}    & / & 94.04 & 95.66 & 93.11   \\
  \hline
  SEGRN (FCN+SEGR) & / & \ding{52}   &  /   &  /&   95.77 &  98.12  &   95.12   \\
  \hline
  FCN+SAGR+SEGR & \ding{52}  & \ding{52}  &    &   & 97.06 &  98.75   &  96.60  \\
  FCN+SAGR+SEGR & \ding{52}  & \ding{52}  &    & \ding{52}  & 97.37 &  98.74  &    96.95 \\
  SSGRN (FCN+SAGR+SEGR+AB) & \ding{52}  & \ding{52}  &  \ding{52} &  \ding{52} &  97.87 &  98.77  &   97.52  \\
  \hline
\end{tabular}
%}
\end{table*}

\begin{table}[t]
  \scriptsize
  \caption{Model complexities in different combinations of the proposed methods on the Indian Pines dataset}
  \newcommand{\tabincell}[2]{\begin{tabular}{@{}#1@{}}#2\end{tabular}}
  \centering
  \label{complexity}
  \resizebox{\linewidth}{!}{
    \begin{tabular}{l|ccccc}
  \hline
  Method & Params(M) & FLOPs(G) & OA(\%) & AA(\%) & Kappa(\%) \\
  \hline
  FCN &  0.49  &  8.70   &  91.06 &  96.35 &  89.71  \\
  \hline
  FCN+PAM & 0.86   &20.31  & 93.95  & 97.06 &  93.01 \\%12.58+7.73
  FCN+RCCA($R=2$) & 0.86   & 16.00 & 93.40  & 95.38 &  92.38 \\%13.44+0.86+1.70
  FCN+SAGR & 0.98  & 12.46 & 89.87 & 91.81  &   88.31  \\%12.00+0.46
  SAGRN (FCN+SAGR+AB)  &  1.28  & 15.54 &  94.04 & 95.66 & 93.11 \\% AB:3.08
  \hline
  FCN+CAM & 0.78   & 12.42 & 94.34 &  97.52  &  93.47 \\%11.74+0.68
  SEGRN (FCN+SEGR) & 1.09  & 12.00 & 95.77 &  98.12  &  95.12 \\%11.9+0.10
  \hline
\end{tabular}
}
\end{table}

\subsection{Parameter Analysis}

In the proposed methods, only the descriptor numbers in SAGRN and SEGRN need to be configured manually. They are exactly the number of nodes in spatial and spectral graphs. They also implicitly determine the average size of homogeneous regions or channels, which affect the scale of context aggregations when generating descriptors. Because of this, we analyze the influences of different descriptor numbers on the network performance, and the corresponding accuracies including OA, AA, and Kappa have been displayed in Figure \ref{k_acc}. It can be observed that the OA keeps increasing as the descriptors grow. Concretely, when increasing descriptors, accuracies are improved quickly in the early stage, but growth rates gradually slow down, and the turning points happen on 16 and 32 in SAGRN and SEGRN, respectively. However, when the descriptor number of SAGRN is larger than 256, accuracies have a slight decline. In our consideration, too manly homogeneous areas fragment the feature, and these small areas narrow the scope of region aggregation, which may weaken the representative ability of the generated descriptors. When there are few descriptors, accuracies may decrease instead even if the descriptor numbers are improved, especially in SEGRN. Moreover, in SEGRN, we also notice that accuracies keep at a low level when the descriptor number is less than 16 because spectral information is lost when a large number of bands are aggregated at once. In our experiments, to obtain relatively high accuracies while reducing the computational complexity as much as possible, we set the descriptor number in SAGRN and SEGRN both to 256.

\subsection{Ablation Study}

We evaluate different parts of the proposed methods on the Indian Pines dataset, including the branches of spatial or spectral graph reasoning (SAGR or SEGR) and the AB in SAGRN. In addition, the influence of multiple loss functions (we mainly consider the additional $L_{sa}$ and $L_{se}$, while $L_{fused}$ is not included) are also considered, and the results are shown in Table \ref{ablation}. It can be seen that the proposed methods improve FCN a lot with the help of graph reasoning. However, without the AB, the accuracies of SAGRN degrade seriously since a rough feature cannot produce high-quality homogeneous areas and effective descriptors. The extra-added AB directly supervises the corresponding feature rather than the reprojected feature like the SAGR path, shortening the distance from feature $F$ to the classifier end, overcoming the difficult learning phenomenon that is caused by weak gradient propagation, and further promoting the network convergence. Thus, the abilities of SAGR are helped to be more fully exploited with the high-quality feature generated under the aid of AB, obtaining better results. Compared with FCN, SEGRN performs better since the contexts lying in different channels are well extracted by implementing SEGR, showing the importance of spectral information in HSI interpretation. It is also a critical characteristic to distinguish HSI from natural images. By jointly taking the advantage of SAGR and SEGR, SSGRN achieves the best performance, especially when equipped with $L_{sa}$ and $L_{se}$, since the gradients gain better propagations in the corresponding SAGRN and SEGRN.

\subsection{Model Complexity}

To more comprehensively analyze the proposed methods, we assess the parameter number (Params) and computational complexity of SAGRN and SEGRN, while the latter is shown in the form of floating-point operations per second (FLOPs), and the results are shown in Table \ref{complexity}. Based on the backbone network of FCN, we simultaneously compare performances of the proposed SAGR and SEGR with other existing commonly employed non-local context capturing modules that utilize self-attention mechanisms, including PAM\cite{danet}, RCCA\cite{ccnet} and CAM\cite{danet}, where the recurrent number $R$ of RCCA is set to 2 to keep consistent with the original literature. It can be seen that our methods perform better than other modules since they achieve competitive accuracies with fewer computations. Concretely, on the square Indian Pines dataset, the inner product operations in PAM and RCCA are separately implemented $N^2$ and $4N\sqrt{N}$ times, while the proposed SAGR only needs $K^2+NK$ times, where $N$ is the pixel number of input features, $K$ is the number of spatial descriptors, in practice $K\ll N$. In SEGRN, the complexity of SEGR is less than CAM since the feature $F$ is downsampled before implementing graph reasoning. Although the accuracies of FCN+PAM on the Indian Pines dataset are close to that of FCN+SAGR+AB, due to memory issues, it may not be able to handle some large scenes at once, such as the University of Houston dataset, which unavoidably affects practical applications. Actually, in our experiments, the combination of FCN, SAGR, and AB equals SAGRN, while FCN+SEGR equals SEGRN.

\subsection{Performance Comparison}

\begin{table}[t]
  \scriptsize
  \newcommand{\tabincell}[2]{\begin{tabular}{@{}#1@{}}#2\end{tabular}}
  \caption{Accuracies of different algorithms on the Indian Pines dataset (\%)}
  \label{indian_table}
  \centering
  %\resizebox{\textwidth}{!}{
  \begin{tabular}{l|ccc}
  \hline
  Method & OA & AA & Kappa \\
  \hline
  1-DCNN & 73.19$\pm$3.61 &83.29$\pm$2.12& 69.52$\pm$3.74\\
  2-DCNN& 82.94$\pm$1.84& 92.88$\pm$0.83 & 80.55$\pm$2.06\\
  SSFCN & 89.62$\pm$1.45 & 94.25$\pm$0.90 & 88.12$\pm$1.65\\
  GCN & 70.22$\pm$1.51 & 79.00$\pm$1.20 & 66.27$\pm$1.53 \\
  Mini-GCN & 76.15$\pm$2.94 & 75.96$\pm$1.79 & 73.11$\pm$3.09\\
  MDGCN & 94.37$\pm$0.67 & 92.91$\pm$1.42 & 93.50$\pm$0.77 \\
  CADGCN& 89.36$\pm$0.82 & 93.24$\pm$1.01 &87.71$\pm$0.93 \\
  DASGCN & 74.84$\pm$1.87 & 76.28$\pm$3.25 & 71.12$\pm$2.19\\
  EMS-GCN & 67.72$\pm$3.29 & 68.37$\pm$3.40 & 63.35$\pm$3.27\\
  \hline
  SAGRN & 94.04$\pm$1.63 & 95.66$\pm$1.99 & 93.11$\pm$1.88 \\
  SEGRN & 95.77$\pm$0.74 & 98.12$\pm$0.36 & 95.12$\pm$0.85 \\
  SSGRN & \bfseries 97.87$\pm$0.55& \bfseries 98.77$\pm$0.19 &\bfseries  97.52$\pm$0.63 \\
\hline
\end{tabular}
 % }
\end{table}

\begin{table}[t]
  \scriptsize
  \newcommand{\tabincell}[2]{\begin{tabular}{@{}#1@{}}#2\end{tabular}}
  \caption{Accuracies of different algorithms on the Pavia University dataset (\%)}
  \label{pavia_table}
  \centering
  %\resizebox{\textwidth}{!}{
  \begin{tabular}{l|ccc}
  \hline
  Method & OA & AA & Kappa \\
  \hline
  1-DCNN & 75.39$\pm$2.49 & 83.13$\pm$1.10 & 68.73$\pm$2.60 \\
  2-DCNN & 88.19$\pm$1.01 & 89.98$\pm$0.53 & 84.49$\pm$1.27 \\
  SSFCN & 85.99$\pm$1.76 & 90.32$\pm$0.95 & 81.83$\pm$2.21 \\
  GCN & 81.83$\pm$2.26 & 87.60$\pm$0.48 & 76.75$\pm$2.62 \\
  Mini-GCN & 85.22$\pm$2.44 & 87.15$\pm$1.23  & 80.74$\pm$2.96 \\
  MDGCN & 95.70$\pm$1.02 & 93.43$\pm$0.44 & 94.30$\pm$1.33 \\
  CADGCN & 92.08$\pm$2.66 &  90.43$\pm$3.36  & 89.64$\pm$3.27 \\
  DASGCN & 89.77$\pm$1.26 & 84.13$\pm$2.15 & 86.30$\pm$1.67\\
  EMS-GCN & 77.69$\pm$3.50 & 83.77$\pm$3.06 & 71.87$\pm$4.29\\
  \hline
  SAGRN & 95.38$\pm$1.33 & 94.44$\pm$1.97 & 93.86$\pm$1.77 \\
  SEGRN & 86.13$\pm$2.48 & 87.49$\pm$1.44 & 81.93$\pm$3.09 \\
  SSGRN &\bfseries 98.70$\pm$0.42 &\bfseries 98.84$\pm$0.31 &\bfseries 98.27$\pm$0.56\\
\hline
\end{tabular}
%  }
\end{table}

\begin{table}[t]
  \scriptsize
  \newcommand{\tabincell}[2]{\begin{tabular}{@{}#1@{}}#2\end{tabular}}
  \caption{Accuracies of different algorithms on the Salinas Valley dataset (\%)}
  \label{salina_table}
  \centering
  %\resizebox{\textwidth}{!}{
  \begin{tabular}{l|ccc}
  \hline
  Method & OA & AA & Kappa \\
  \hline
  1-DCNN & 86.92$\pm$3.82 & 93.63$\pm$1.09 & 85.49$\pm$4.12\\
  2-DCNN & 87.58$\pm$2.33 & 92.87$\pm$1.08 & 86.21$\pm$2.55 \\
  SSFCN & 92.99$\pm$0.35 & 96.49$\pm$0.37 & 92.20$\pm$0.40\\
  GCN & 88.21$\pm$1.07 & 94.14$\pm$0.46 & 86.86$\pm$1.18 \\
  Mini-GCN & 90.12$\pm$0.14 & 93.92$\pm$0.37 & 89.00$\pm$0.14\\
  MDGCN & 98.37$\pm$0.31 & 98.46$\pm$0.16 & 98.18$\pm$0.35 \\
  CADGCN & 97.93$\pm$0.29 & 98.11$\pm$0.38 & 97.69$\pm$0.32\\
  DASGCN & 92.64$\pm$1.27 & 92.40$\pm$1.71 & 91.81$\pm$1.42\\
  EMS-GCN & 93.31$\pm$1.17 & 92.90$\pm$1.56 & 92.55$\pm$1.29\\
  \hline
  SAGRN & 96.91$\pm$0.58 & 98.48$\pm$0.33 & 96.55$\pm$0.65 \\
  SEGRN & 92.22$\pm$2.77 & 93.68$\pm$2.24 & 91.36$\pm$3.06 \\
  SSGRN & \bfseries 99.34$\pm$0.46 & \bfseries 99.60$\pm$0.21 & \bfseries 99.26$\pm$0.51\\
\hline
\end{tabular}
%  }
\end{table}

\begin{table}[t]
  \scriptsize
  \newcommand{\tabincell}[2]{\begin{tabular}{@{}#1@{}}#2\end{tabular}}
  \caption{Accuracies of different algorithms on the University of Houston dataset (\%)}
  \label{houston_table}
  \centering
  %\resizebox{\linewidth}{!}{
  \begin{tabular}{l|ccc}
  \hline
  Method & OA & AA & Kappa \\
  \hline
  1-DCNN & 80.23$\pm$0.84 & 82.09$\pm$0.89 & 78.61$\pm$0.91 \\
  2-DCNN & 84.08$\pm$1.18 & 87.01$\pm$0.93 & 82.77$\pm$1.28 \\
  SSFCN & 84.14$\pm$3.26 &85.84$\pm$2.98 & 82.86$\pm$3.51 \\
  GCN&  85.73$\pm$0.95 &86.41$\pm$0.79 & 84.54$\pm$1.03 \\
  Mini-GCN &  83.51$\pm$1.86& 84.46$\pm$1.66 & 82.15$\pm$2.01 \\
  MDGCN & 92.13$\pm$0.41 & 93.25$\pm$0.33 & 91.48$\pm$0.45 \\
  CADGCN&93.05$\pm$0.68 &93.27$\pm$0.54& 92.48$\pm$0.74\\
  DASGCN & 67.65$\pm$4.14 & 70.51$\pm$4.34 & 65.02$\pm$4.49\\
  EMS-GCN & 61.87$\pm$1.47 & 65.22$\pm$1.56 & 58.89$\pm$1.60\\
  \hline
  SAGRN &  92.38$\pm$1.94 & 92.20$\pm$1.83 &91.75$\pm$2.10 \\
  SEGRN &  89.87$\pm$1.52 & 91.80$\pm$1.24 &  89.04$\pm$1.63 \\ 
  SSGRN &\bfseries 95.59$\pm$1.95 &\bfseries 96.52$\pm$1.56&\bfseries 95.23$\pm$2.11 \\
\hline
\end{tabular}
%  }
\end{table}

We conduct a comparison between the proposed methods with other classical or state-of-the-art approaches, including typical CNN-based networks: 1-DCNN \cite{3dcnn} for spectral classification with spectral vector, 2-DCNN \cite{3dcnn} for spatial classification using spatial patches, and an FCN-based algorithm SSFCN \cite{ssfcn} for spectral-spatial joint classification. As for approaches on the foundation of graph convolution, we choose classical GCN \cite{kipfgcn} and Mini-GCN \cite{minigcn}, while MDGCN\cite{mdgcn}, CADGCN \cite{cadgcn}, DASGCN \cite{dasgcn} and EMS-GCN \cite{emsgcn} are selected as advanced GCN relevant techniques. The implementation of GCN is in the pattern of transductive learning, where training and testing sets are simultaneously fed into the model since each pixel sample is regarded as a graph node, and it unavoidably consumes too many computational resources. Mini-GCN simplifies this problem by using a shrunk adjacency matrix that is obtained through only computing on the nodes in the current mini-batch, while the graph nodes in MDGCN are obtained from the segmented superpixels and the classification is performed at the superpixel-level, too. CADGCN also uses superpixels to generate graph nodes and performs a pixel-region-pixel transformation. By using predefined superpixels, DASGCN simultaneously obtains spatial and spectral information and samples different pixels as nodes for performing graph convolutions. While EMS-GCN captures spatial graph contexts by dynamic superpixel maps that are generated by networks, they additionally employ channel attention to enhance spectral features.

\begin{figure*}[t]
  \centering
  \includegraphics[width=\linewidth]{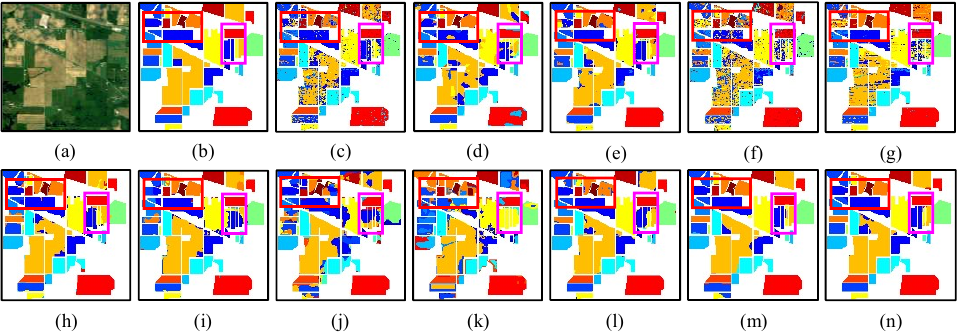}
  \caption{Classification map of different methods on the Indian Pines dataset. (a) Original Image. (b) Ground Truth (c) 1-DCNN (d) 2-DCNN (e) SSFCN. (f) GCN. (g) Mini-GCN. (h) MDGCN. (i) CADGCN. (j) DASGCN. (k) EMS-GCN. (l) SAGRN. (m) SEGRN. (m) SSGRN.}
  \label{indian_map}
\end{figure*}

\begin{figure}[t]
  \centering
  \includegraphics[width=\linewidth]{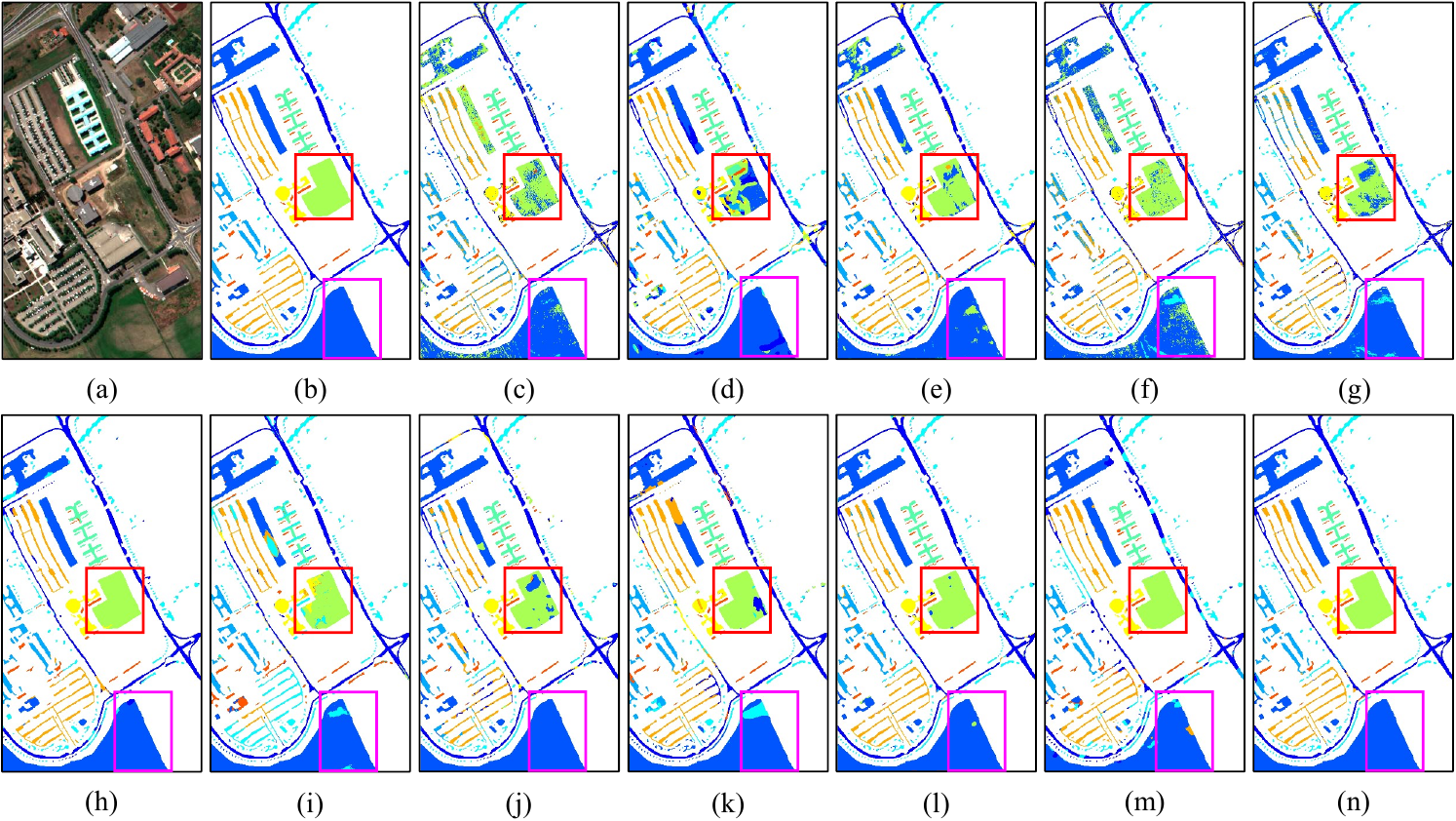}
  \caption{Classification map of different methods on the Pavia University dataset. (a) Original Image. (b) Ground Truth (c) 1-DCNN (d) 2-DCNN (e) SSFCN. (f) GCN. (g) Mini-GCN. (h) MDGCN. (i) CADGCN. (j) DASGCN. (k) EMS-GCN. (l) SAGRN. (m) SEGRN. (m) SSGRN.}
  \label{pavia_map}
\end{figure}

\begin{figure}[t]
  \centering
  \includegraphics[width=\linewidth]{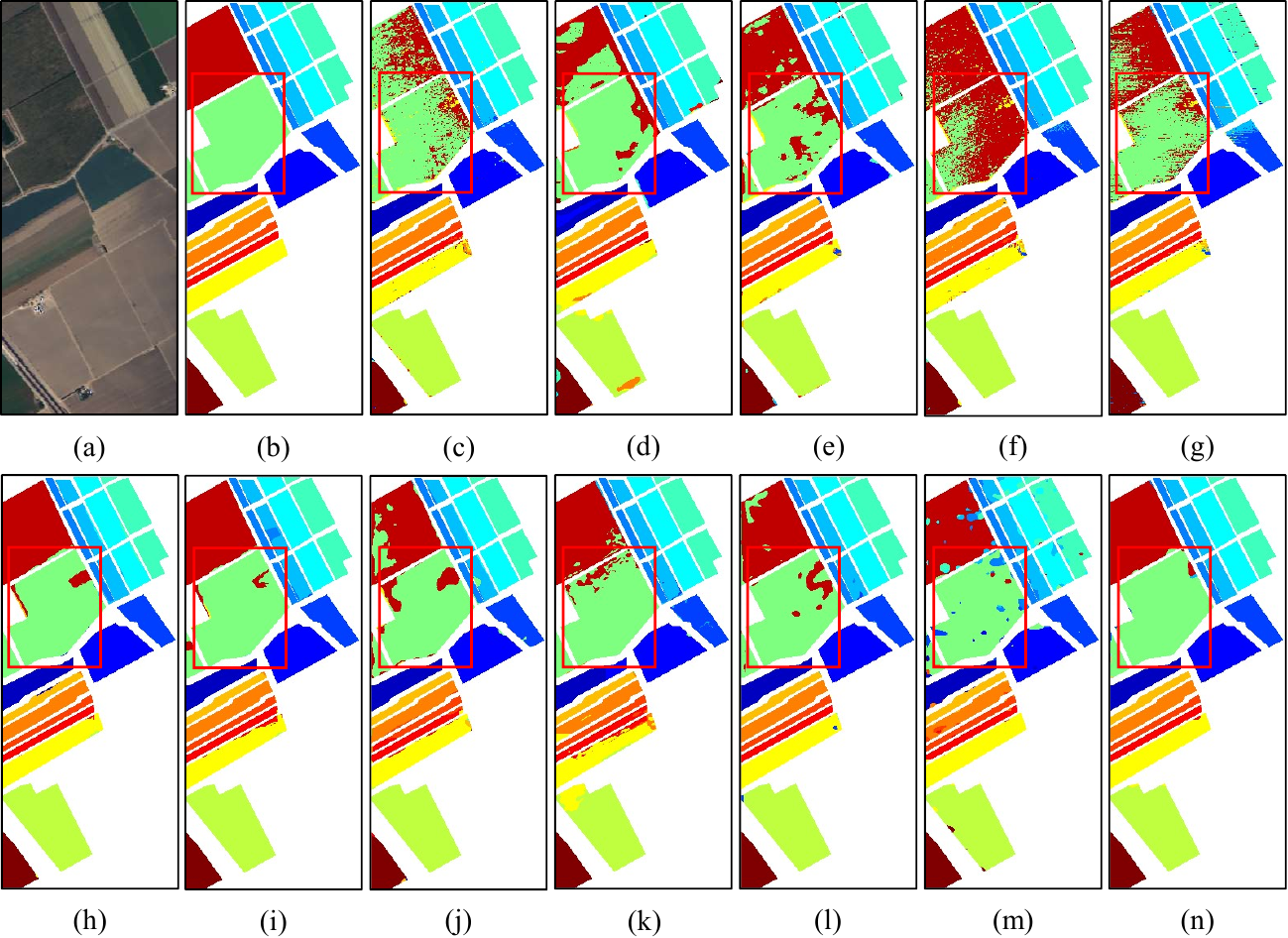}
  \caption{Classification map of different methods on the Salinas Valley dataset. (a) Original Image. (b) Ground Truth (c) 1-DCNN (d) 2-DCNN (e) SSFCN. (f) GCN. (g) Mini-GCN. (h) MDGCN. (i) CADGCN. (j) DASGCN. (k) EMS-GCN. (l) SAGRN. (m) SEGRN. (m) SSGRN.}
  \label{salina_map}
\end{figure}

\begin{figure*}[t]
  \centering
  \includegraphics[width=\linewidth]{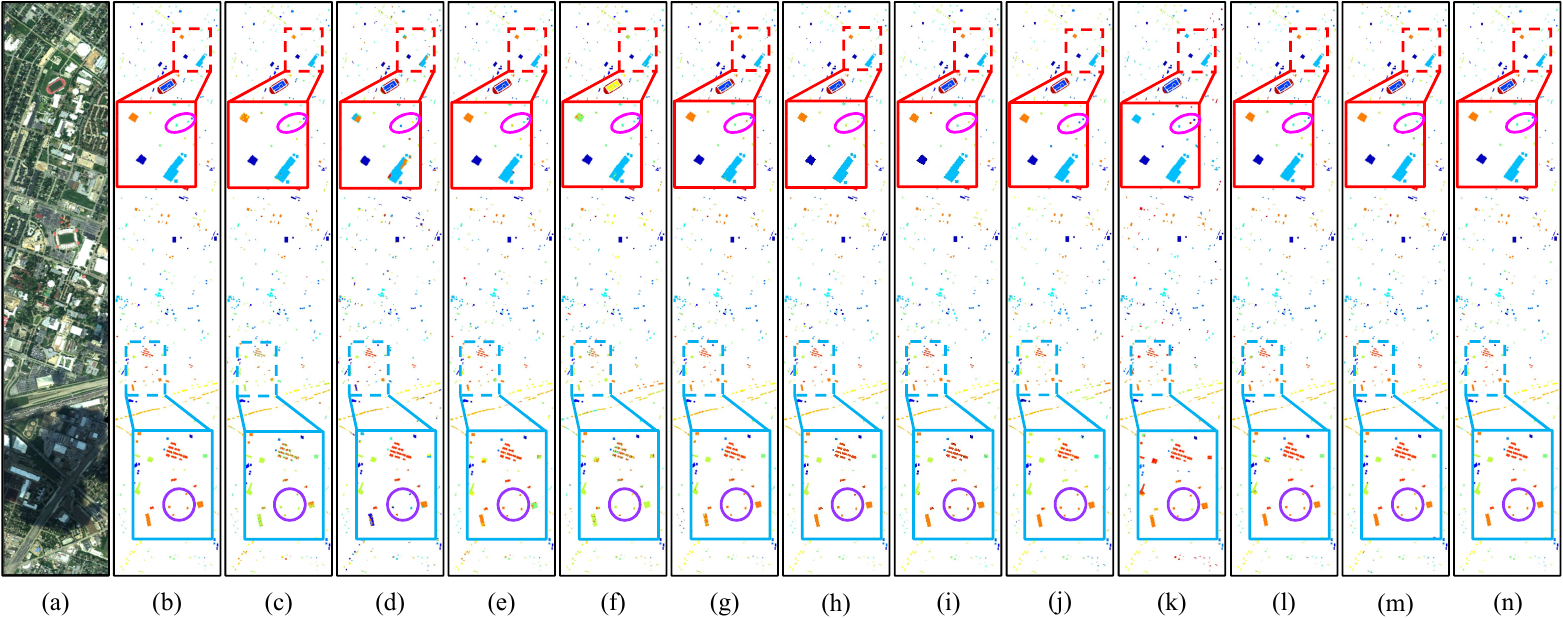}
  \caption{Classification map of different methods on the University of Houston dataset. (a) Original Image. (b) Ground Truth (c) 1-DCNN (d) 2-DCNN (e) SSFCN. (f) GCN. (g) Mini-GCN. (h) MDGCN. (i) CADGCN. (j) DASGCN. (k) EMS-GCN. (l) SAGRN. (m) SEGRN. (m) SSGRN.}
  \label{houston_map}
\end{figure*}

Table \ref{indian_table}-\ref{houston_table} lists the classification accuracies of each algorithm, where the mean value and standard deviation are reported at the same time. It can be seen that the FCN-based SSFCN usually performs better than the CNN-based 1-DCNN and 2-DCNN, showing the importance of spectral-spatial combination in HSIC. However, the accuracies of SSFCN are still limited because SSFCN only perceives local information since its convolutions are local operators, and regular context capturing can not align with irregular object distributions. For graph convolution-based methods, redundant calculations of adjacent pixels in GCN affect the classification while Mini-GCN attempts to use a simplified adjacency matrix to accelerate this procedure with a mini-batch training strategy. However, in GCN and Mini-GCN, pixels without representation of various contexts are directly set as descriptors, degrading the classification. Some effective descriptors are obtained in MDGCN, CADGCN, and DASGCN with the help of superpixel segmentation on the original image. Thus, they get higher accuracies than GCN and Mini-GCN, especially on Pavia University and Salinas Valley datasets. What needs to be noticed is that the adjacency matrices in MDGCN, CADGCN, and DASGCN are both calculated by only considering the relationships between neighbor nodes, which may be limited in the long strip University of Houston scene. Among these methods, DASGCN performs the worst since it conducts the pixel-level graph convolution that has shorter context ranges. We also notice EMS-GCN does not perform well in almost all datasets. It is probably because the obtained dynamic superpixel map during network training is not stable. At this time, the produced graph features by local context aggregations may be not effective. Compared with the above methods, although homogeneous areas are produced by implementing superpixel segmentation on intermediate features in the network, our SAGRN performs better since we adopt a global view where the relationships between each node and all the other nodes are measured, and more effective descriptors can be flexibly and adaptively generated. By implementing spectral reasoning, SEGRN successfully captures the relationships lying in different channels. At last, combining the spatial and spectral graph contexts that are separately obtained by SAGRN and SEGRN, the proposed SSGRN achieves the best overall accuracies of 97.87\%, 98.70\%, 99.34\%, and 95.59\%, respectively, exceeding the second place by 3.50\%, 3.00\%, 1.41\%, and 2.54\% on Indian Pines, Pavia University, Salinas Valley and University of Houston datasets.

\begin{figure}[t]
  \centering
  \subfigure[]{\includegraphics[width=0.48\linewidth]{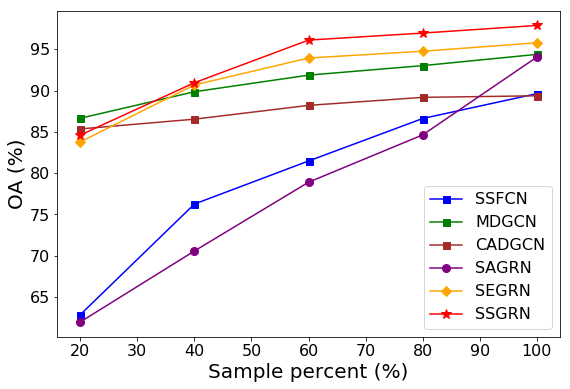}}
  \subfigure[]{\includegraphics[width=0.48\linewidth]{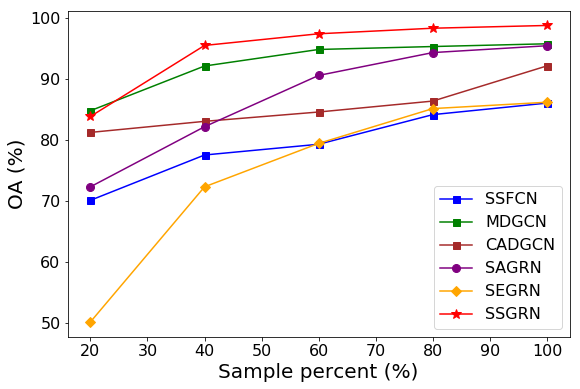}}\\
  \subfigure[]{\includegraphics[width=0.48\linewidth]{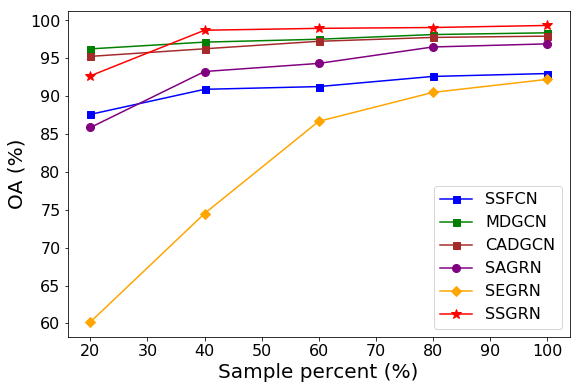}}
  \subfigure[]{\includegraphics[width=0.48\linewidth]{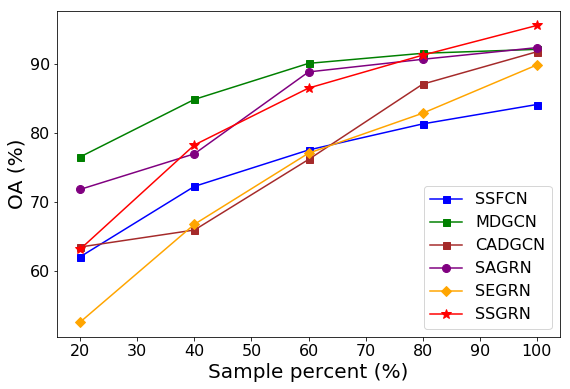}}
  \caption{Relationship between OA and the different number of training samples, where the X axis is the percentage of training samples in Table \ref{indian_dataset_tab}-\ref{houston_dataset_tab}. (a) Indian Pines. (b) Pavia University. (c) Salinas Valley. (d) University of Houston.}
  \label{sample_acc}
\end{figure}

The classification maps of the above methods have been depicted in Figure \ref{indian_map}-\ref{houston_map}. On the Indian Pines dataset, it can be seen that the spectral vector-based methods 1-DCNN, GCN, and Mini-GCN have serious point noises. The introduced spatial information alleviates this issue, making the maps of 2-DCNN and SSFCN cleaner. However, their conventional convolutional filters can not capture irregular contexts, especially in areas where objects are densely distributed, such as red and pink boxes. In GCN-based methods, MDGCN and CADGCN partly solve this problem by adopting convolutions on graph structures. Compared with them, the proposed methods have longer graph context perception distances, and we also consider spectral information. Therefore, SSGRN can simultaneously obtain discriminative results where the objects possess continuous surfaces, well-maintained edges, and are close to the ground truth in the box displayed areas. On the Pavia University dataset, our method produces more smooth surfaces inside objects, see red and pink boxes. In the region of the red box at the Salinas Valley scene, the proposed method generates a preferable visual result with fewer misclassifications compared to other approaches. On the challenging University of Houston dataset, whose classification maps are difficult to be distinguished, since a large number of small objects are distributed dispersedly. Thus, we select and enlarge two typical regions for comparison, which have been presented in red and blue boxes. We can find that MDGCN and the proposed methods perform well on relatively large objects, so we further strengthen restrictions with pink and purple circles, respectively. It can be seen that benefitting from the global graph context capturing, only our SAGRN and SSGRN correctly classify all small objects in these areas. In addition, in our intuition, SEGRN is probably not as good as SAGRN since it only exploits spectral contexts. Surprisingly, SEGRN performs better than SAGRN on the Indian Pines dataset. It may be because the Indian Pines image possesses more channels. This result implies the importance of channel information in HSIC.

\begin{figure}[t]
  \centering
  \subfigure[]{\includegraphics[width=0.24\linewidth]{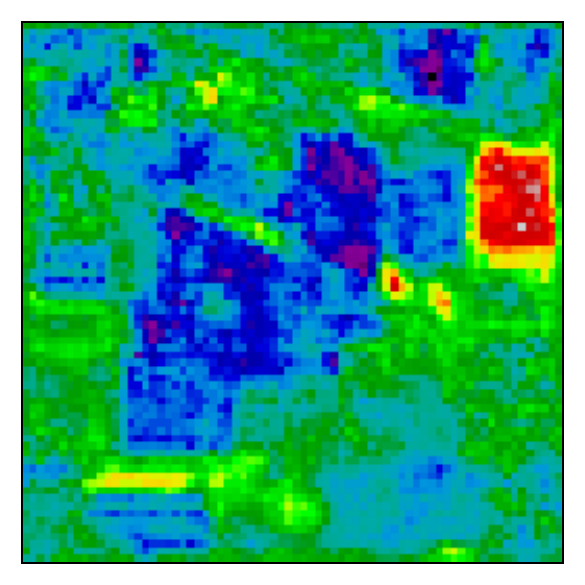}}
  \subfigure[]{\includegraphics[width=0.24\linewidth]{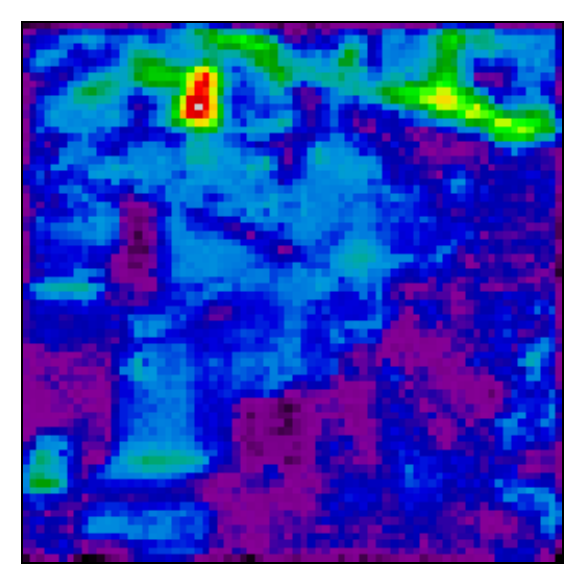}}
  \subfigure[]{\includegraphics[width=0.24\linewidth]{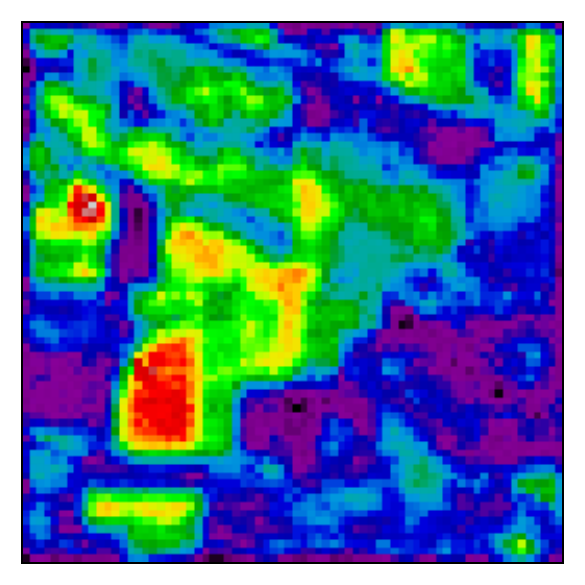}}
  \subfigure[]{\includegraphics[width=0.24\linewidth]{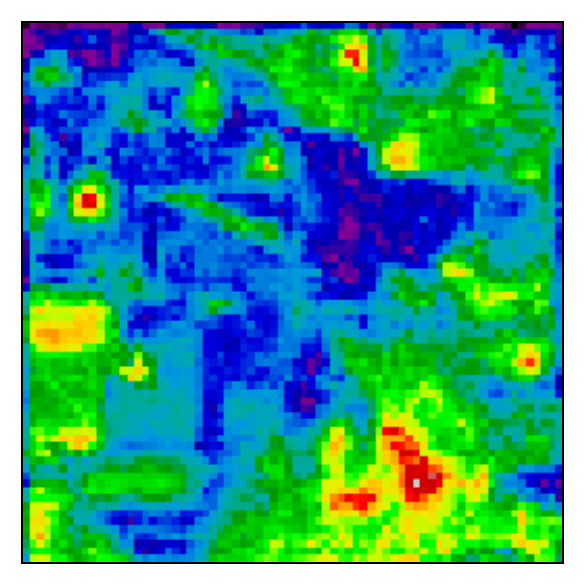}}
  \caption{Responsibility of different descriptors at each spatial position of feature $F$ on the Indian Pines dataset. The red color represents high intensity. (a) Descriptor 1. (b) Descriptor 2. (c) Descriptor 3. (d) Descriptor 4.}
  \label{sa_descriptor}
\end{figure}

\begin{figure}[t]
  \centering
  \includegraphics[width=\linewidth]{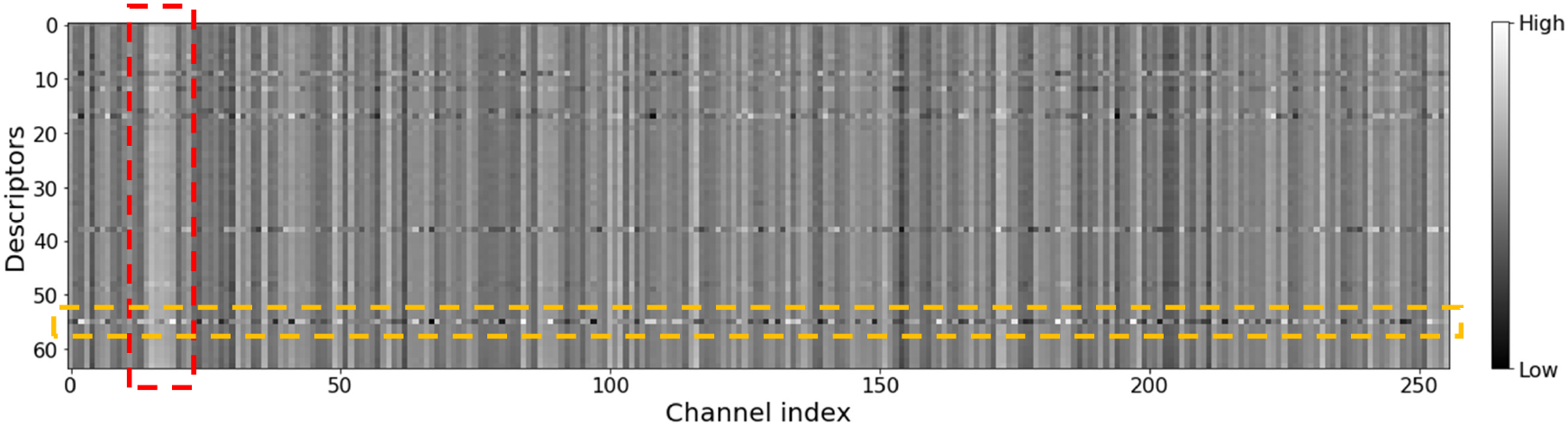}
  \caption{Responsibility of different descriptors at each channel of feature $F$ on the Indian Pines dataset. For convenience, here, the number of spectral descriptors is set to 64.}
  \label{se_descriptor}
\end{figure}

\begin{figure*}[t]
  \centering
  \subfigure[]{\includegraphics[width=0.32\linewidth]{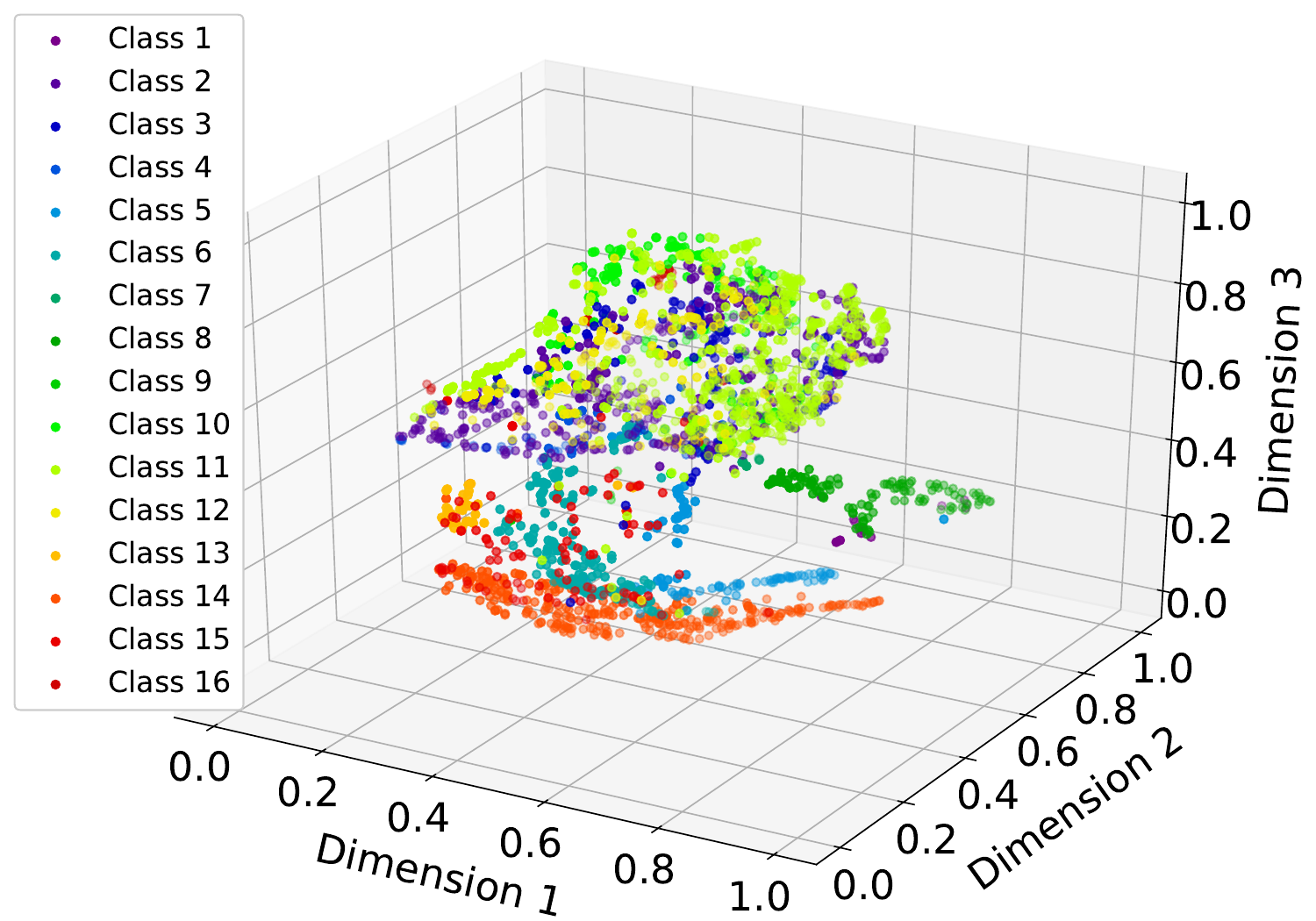}}
  \subfigure[]{\includegraphics[width=0.32\linewidth]{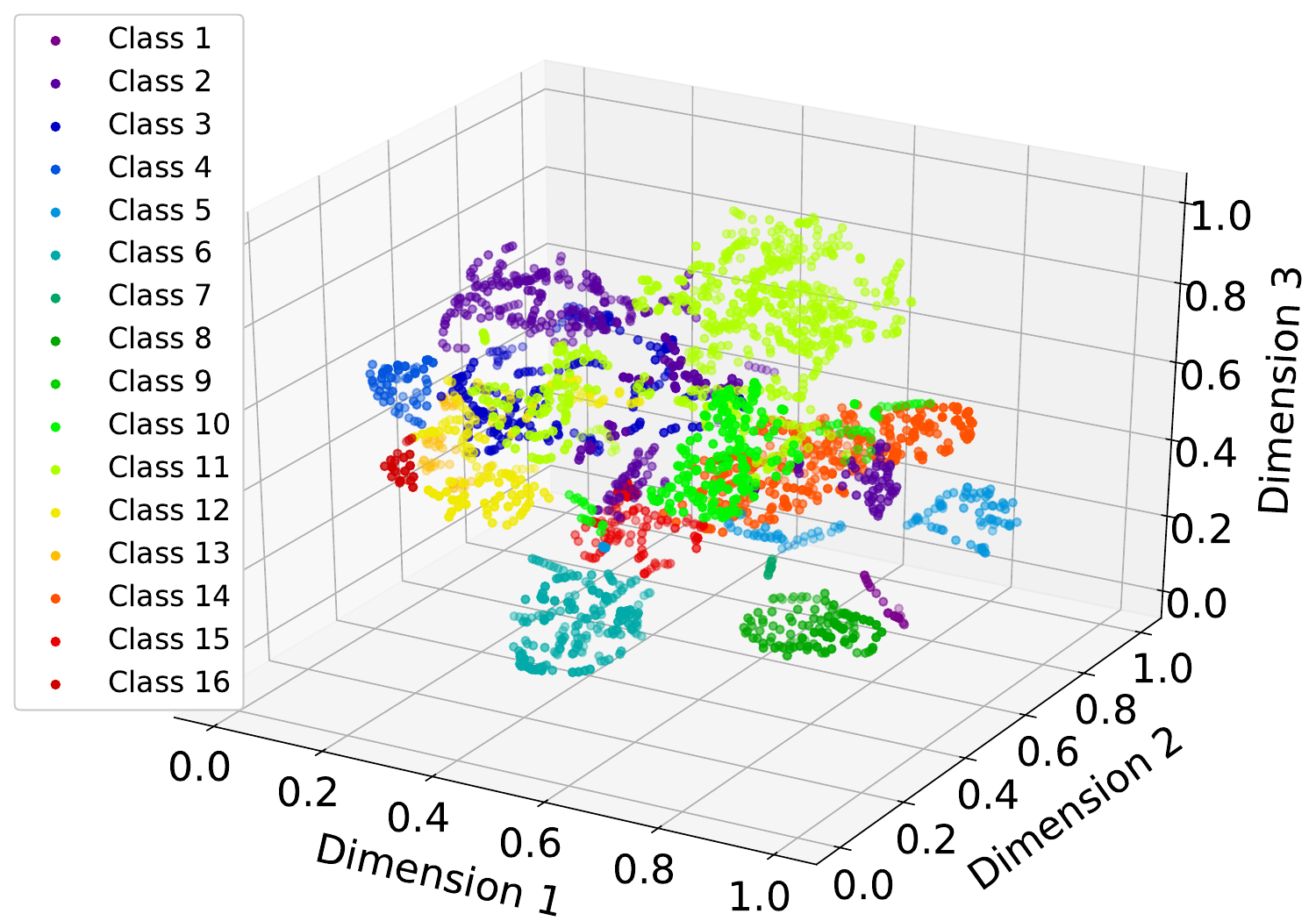}}
  \subfigure[]{\includegraphics[width=0.32\linewidth]{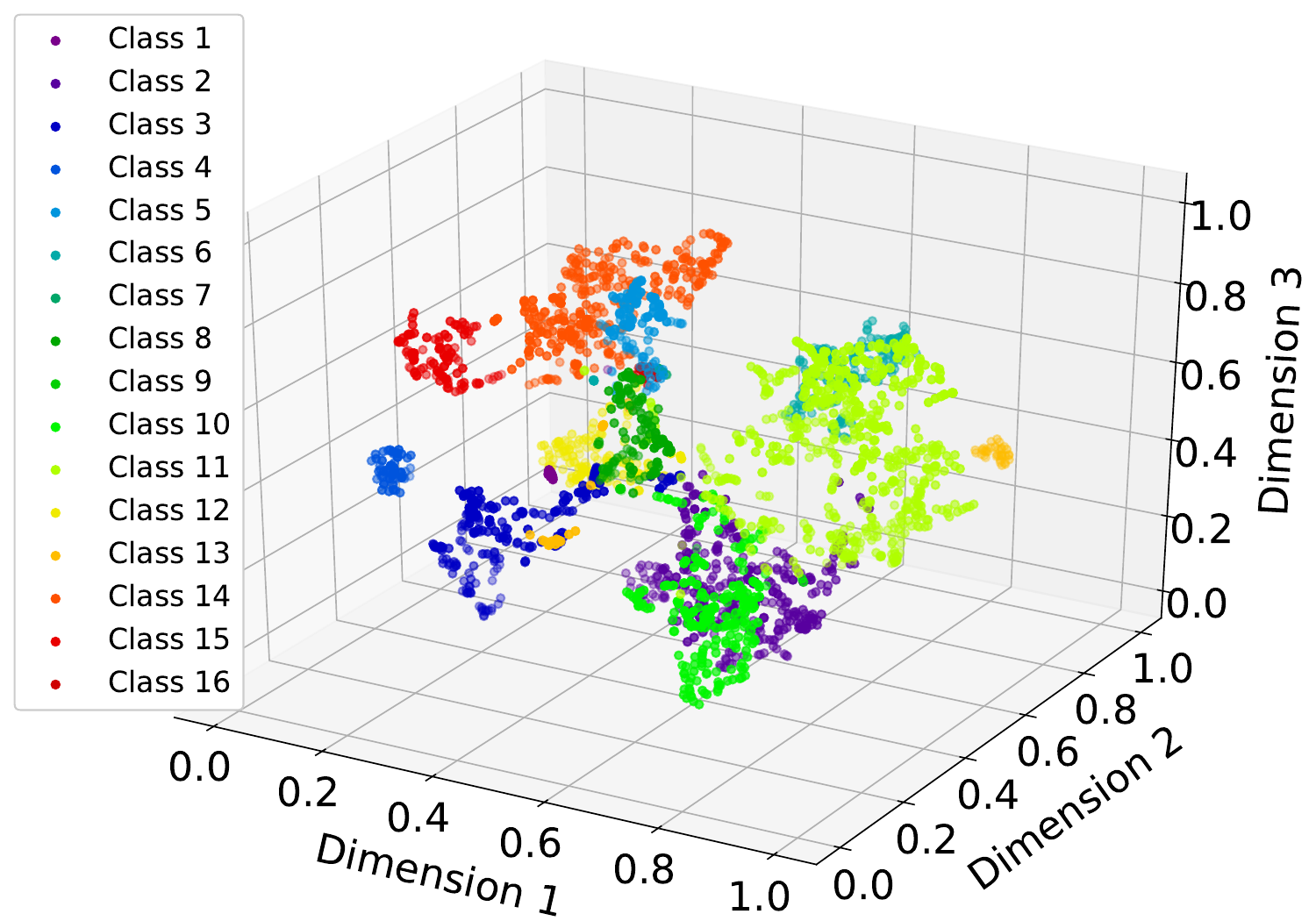}}\\
  \subfigure[]{\includegraphics[width=0.32\linewidth]{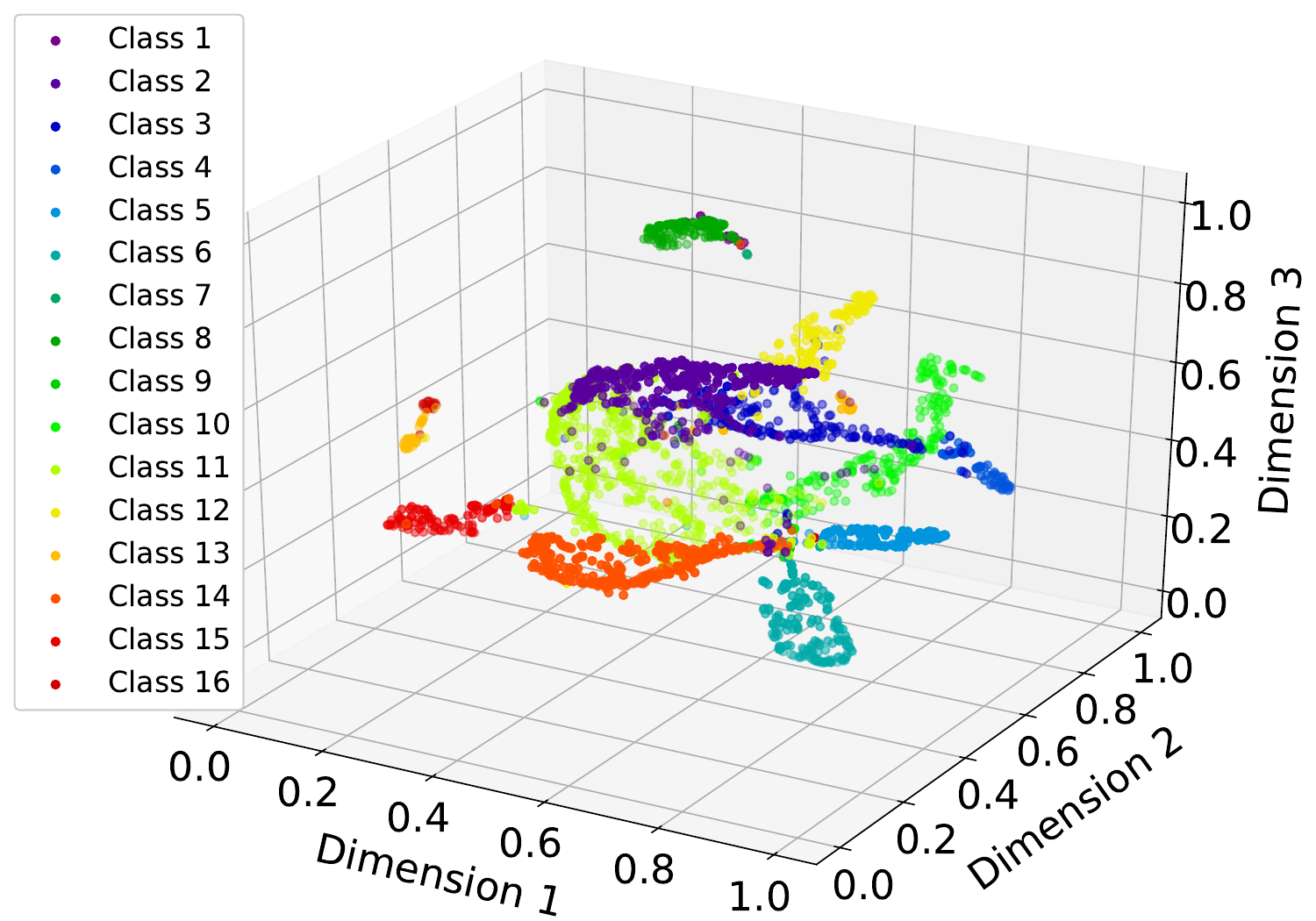}}
  \subfigure[]{\includegraphics[width=0.32\linewidth]{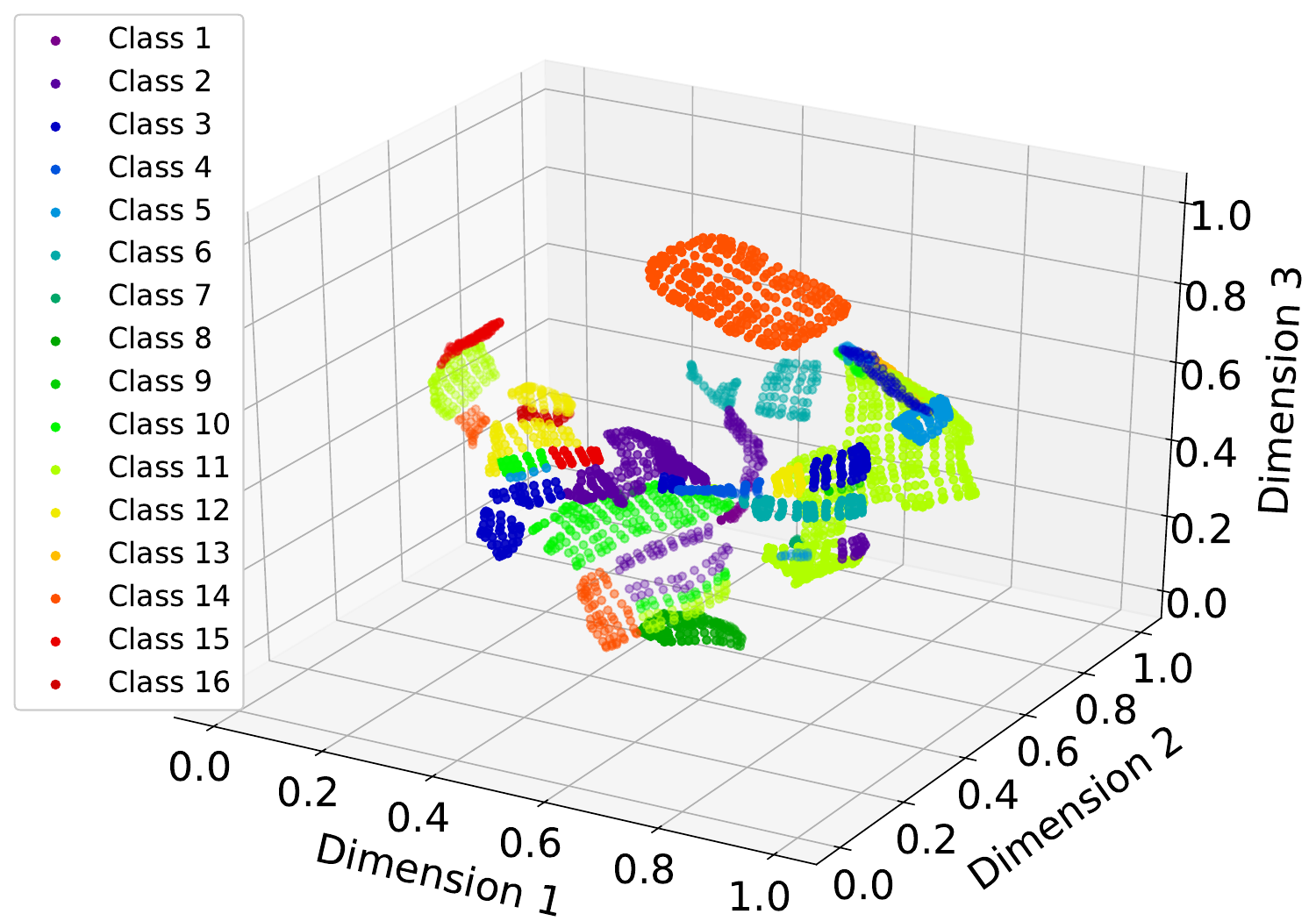}}
  \subfigure[]{\includegraphics[width=0.32\linewidth]{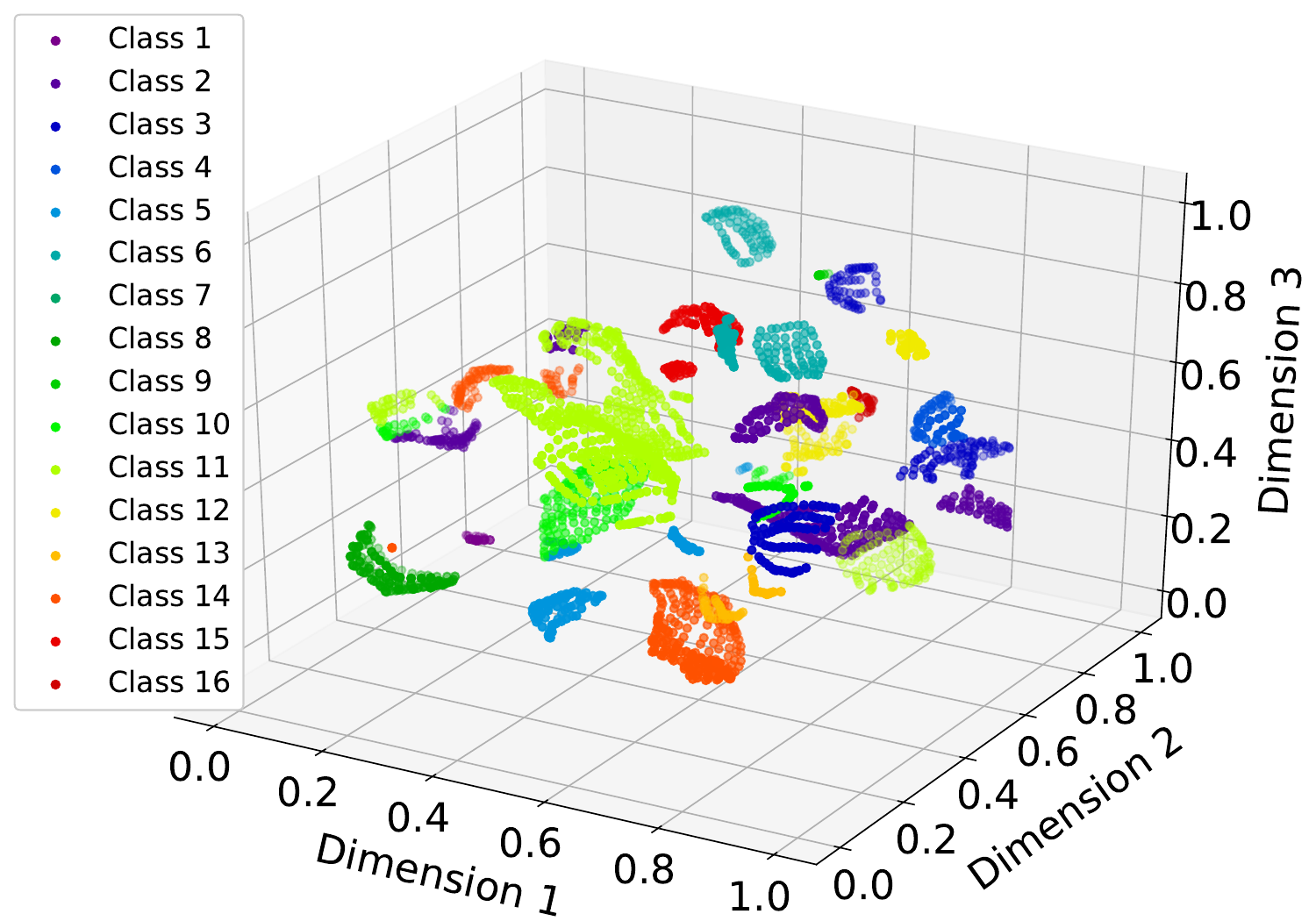}}
  \caption{Distribution of reference samples in different feature spaces on the Indian Pines dataset. (a) Input image. (b) Feature $F$. (c) Probability matrix $P_{sa\_aux}$ in SAGRN auxiliary branch. (d) The extracted feature $F_{sa\_main}$ in SAGRN main branch. (e) The extracted feature $F_{se}$ in SEGRN. (f) The fused feature $F_{fused}$.}
  \label{feature_distribution}
\end{figure*}

We also evaluate the stability of different methods. Specifically, we adopt different percentages for training samples in Table \ref{indian_dataset_tab}-\ref{houston_dataset_tab}. To highlight comparison, we only evaluate the methods that have relatively high accuracies in previous performance comparison experiments, including SSFCN, MDGCN, CADGCN, SAGRN, SEGRN, and SSGRN. The corresponding overall accuracies are shown in Figure \ref{sample_acc}. In an overall view, SSFCN performs the worst and has no advantage in almost all cases since its conventional convolutions cannot perceive irregularly distributed objects. By addressing this problem using graph convolution with the help of superpixel segmentation, MDGCN and CADGCN perform better than SSFCN. In addition, since their superpixels are always generated on the original image, the number of superpixel regions is far less than the pixel number. Thus, the performances of MDGCN and CADGCN are still stable when using fewer samples, especially for MDGCN that adopts superpixel-level prediction. Owing to the proposed graph reasonings are based on high-level features, whose qualities heavily depend on the learnable knowledge of the training data. Thus, our methods may not be as effective as MDGCN and CADGCN when training samples are extremely scarce (e.g. 20\%). Nonetheless, the proposed SAGRN and SEGRN are rapidly improved once the number of training samples is increased. Notice that SEGRN is more preferring the Indian Pines scene, while SAGRN does the opposite. By combining spatial and spectral graph information, SSGRN achieves promising results in most sample scenarios and performs to be comparable even if in cases of few samples.

\subsection{Visualization}

To more intuitively understand our methods, we separately visualize the affinity matrices $A$ in SAGRN and SEGRN in Figure \ref{sa_descriptor}-\ref{se_descriptor}, which separately indicates the similarity of the selected descriptors in different spatial positions or channels of the feature $F$, and the responsibility intensity is represented by different colors. In Figure \ref{sa_descriptor}, the red color means high affinities. It can be seen that different descriptors highlight different areas, demonstrating that they separately possess closer relationships with corresponding regions. In other words, the connotations of these descriptors are certainly the meaning of emphasized areas. These concepts serve as basic components that can be organized by linearly aggregating the transformed graph nodes in $G$ to generate more complex semantic information for understanding other positions. In Figure \ref{se_descriptor}, each row represents the weights on different channels for a specific descriptor, while the columns represent different channels. If one descriptor pays more attention to some channels, then the colors of these columns will be closer to white. To our surprise, most descriptors present similar characteristics. For example, almost all descriptors show high responsibilities on channels in the red box. Nevertheless, there are also some descriptors behaving differently and bringing diverse spectral information, which can be exemplified by the descriptor in the orange box. Although only a few descriptors perform in special, SEGRN still achieves comparable accuracies. The visualization of spectral descriptors also indicates spectral features still have great potentials to be mined for HSIC.

We also assess the distinguishability of extracted features in the proposed SSGRN at a three-dimensional space by utilizing t-SNE dimension reduction \cite{tsne}. The distributions for reference samples at the corresponding input image, the feature $F$ generated by the backbone network, the probability matrix $P_{sa\_aux}$ in AB, the $F_{sa\_main}$ of SAGRN, the $F_{se}$ of SEGRN, and the fused feature $F_{fused}$ are separately shown in Figure \ref{feature_distribution}. It can be observed that the samples on the original image space are mixed-up and difficult to be identified, while after backbone network encoding, the obtained $F$ starts to be separable, and AB further strengthens the distinction. Benefitting from graph reasoning modules, the pixel representations of $F_{sa\_main}$, $F_{se}$ and $F_{fused}$ possess high separability with larger inter-class distances, and the points of each category constitute a unique manifold, indicating that the latent patterns of corresponding categories have been perceived, demonstrating the effectiveness of the proposed SSGRN.

\section{Conclusion}

In this paper, we propose a network called SSGRN to classify the HSI. Considering the irregular distributions of land objects and the various relationships among different spectral bands, the corresponding contextual information is more suitable to be extracted from a graph perspective. Concretely, this network contains two subnetworks that separately extract spatial and spectral graph contexts. In spatial subnetwork SAGRN, to generate more effective descriptors for graph reasoning, different from previous approaches implementing superpixel segmentation on the original image, we move this procedure to intermediate features inside the network. Based on pixel spectral-spatial similarities, we can flexibly and adaptively produce homogeneous regions. Then, descriptors are acquired by separately aggregating these regions. In addition, we conduct a similar operation on channels to obtain spectral graph contexts in a spectral subnetwork named SEGRN, where the spectral descriptors are gained by reasonably grouping different channels. These graph reasoning procedures in spectral and spatial subnetworks are all achieved with graph convolution, where the adjacent matrices are obtained by computing the similarities among all nodes to ensure global perceptions. SSGRN is finally produced by combining SAGRN and SEGRN to further improve the classification. It needs to be noticed that auxiliary branch and multiple loss strategy are separately applied to SAGRN and SSGRN to accelerate network convergence. A series of extensive experiments show that the proposed methods not only maintain high accuracy but also reduce computational resource consumption. Quantitative and qualitative performance comparisons show the competitiveness of the proposed methods compared with other state-of-the-art approaches, even if in the case of fewer samples. The final visualizations make the proposed methods more convincing.

Since the descriptors are obtained by directly aggregating the surrounding areas or channels of the target pixel or band, which may contain irrelevant information and bring noises. Thus, in future work, we will improve the descriptor generation procedure to obtain more serviceable representations.

\bibliographystyle{IEEEtran}
\bibliography{HSI_SSGRN}

\end{document}